\documentclass[times, review, 10pt]{elsarticle}

\usepackage[numbers]{natbib}
\usepackage{setspace}
\usepackage{float}
\usepackage{subcaption}

\usepackage{amssymb}
\usepackage{amsmath}
\usepackage{latexsym}
\usepackage{gensymb}

\usepackage{url} 
\usepackage{verbatim} 
\usepackage{multirow}

\usepackage[colorinlistoftodos]{todonotes}
\usepackage{paralist}

\def\tsc#1{\csdef{#1}{\textsc{\lowercase{#1}}\xspace}}
\tsc{WGM}
\tsc{QE}
\tsc{EP}
\tsc{PMS}
\tsc{BEC}
\tsc{DE}

\begin{document}

\begin{frontmatter}



\title{
evTransFER: A Transfer Learning Framework for Event-based Facial Expression Recognition
}



%

\author[a]{Rodrigo Verschae\corref{c}}
\ead{rodrigo@verschae.org}
\author[a]{Ignacio Bugueno-Cordova}
\cortext[c]{Corresponding author}
\address[a]{Institute of Engineering Sciences, Universidad de O'Higgins, Libertador Bernardo O'Higgins 611, Rancagua, O'Higgins, 2841959, Chile}

\begin{abstract}
Event-based cameras are bio-inspired sensors that asynchronously capture pixel intensity changes with microsecond latency, high temporal resolution, and high dynamic range, providing information on the spatiotemporal dynamics of a scene. We propose evTransFER, a transfer learning-based framework for facial expression recognition using event-based cameras. The main contribution is a feature extractor designed to encode facial spatiotemporal dynamics, built by training an adversarial generative method on facial reconstruction and transferring the encoder weights to the facial expression recognition system. We demonstrate that the proposed transfer learning method improves facial expression recognition compared to training a network from scratch. We propose an architecture that incorporates an LSTM to capture longer-term facial expression dynamics and introduces a new event-based representation called TIE. We evaluated the framework using both the synthetic event-based facial expression database e-CK+ and the real neuromorphic dataset NEFER. On e-CK+, evTransFER achieved a recognition rate of 93.6\%, surpassing state-of-the-art methods. For NEFER, which comprises event sequence with real sensor noise and sparse activity, the proposed transfer learning strategy achieved an accuracy of up to 76.7\%. In both datasets, the outcomes surpassed current methodologies and exceeded results when compared with models trained from scratch.
\end{abstract}

\begin{keyword}
Event-based Cameras \sep Transfer Learning\sep Deep Learning \sep Facial Expression Recognition \sep Facial Reconstruction 
\end{keyword}

\end{frontmatter}


\section{Introduction}
\label{sec:intro}

Event-based neuromorphic vision sensors were introduced by \cite{lichtsteiner2008}, inspired by the asynchronous and sparse processing of biological vision systems.
These event-based cameras are distinguished by their response to local intensity changes in a scene, generating an asynchronous event sequence that captures the timestamp and polarity associated with the brightness change of each pixel. 
Recent surveys on event-based cameras and their applications can be found in \cite{Gallego2019-wf, zheng2023deeplearningeventbasedvision, iddrisu2024eventeyemotionsurvey, chakravarthisurvey2025}.

Due to these novel vision sensors, many works have proposed event-data representations and methods targeted for this device(see e.g. \cite{gehrig2019est, lin2020sae, innocenti2021tbr}). This includes adapting traditional vision methods to event-based tasks, such as tracking \cite{dong2021, mondal2021}, image reconstruction \cite{rebecq2019highspeed, zou2021learningtoreconstruct}, and pattern recognition \cite{chen2020dynamicgraphcnn, kim2021nimagenet}. Nevertheless, there is room for improvement in complex tasks, such as facial expression, where learning to handle short- and long-term spatiotemporal dynamics is essential. Moreover, few annotated datasets exist for event-based cameras; therefore, building accurate systems that do not rely on large, manually annotated datasets remains highly relevant.

To address these issues, we present evTransFER, a new learning framework for event-based facial expression recognition. This learning framework employs transfer learning to train an encoder that captures spatiotemporal dynamics in faces, incorporating a new event data representation to enhance recognition. To evaluate evTransFER, we use an event-based facial expression database called e-CK+ \cite{verschae2023eventgesturefer} and the NEFER dataset \cite{berlincioni2023nefer}. The proposed method outperforms current event-based methods on these datasets (one simulated and one real), particularly when compared with methods for similar problems. The project website, featuring additional information and materials, can be found at: \url{https://uoh-rislab.github.io/evtransfer}.

The remainder of this paper is organized as follows. 
Section~\ref{sec:state-of-the-art} presents the state-of-the-art methods associated with event-based cameras and facial analysis. Section~\ref{sec:preliminaries} describes representations, classification methods, available datasets, and tools for event-based cameras.
Section~\ref{sec:methodology} presents the proposed event-based facial expression recognition framework, evTransFER. 
Section~\ref{sec:evaluationSetUp} describes the experimental setup and
Section~\ref{sec:results} presents the evaluation results. Finally, Section~\ref{sec:discussion} discusses these results, and Section~\ref{sec:conclusions} concludes and outlines future research.

\section{Related Work}
\label{sec:state-of-the-art}

Face analysis is a well-studied area in computer vision, using classical and deep learning-based techniques to solve various problems \cite{zhao2003frsurvey, jafri2009surveyfr, wang2021dfrsurvey}. For event-based facial analysis, several works report on neuromorphic face analysis \cite{becattini2025neuromorphicfaceanalysissurvey}, including face detection, eye blinking detection, and facial expression recognition, alongside tasks like (i)drowsiness-driving detection \cite{chen2020eddd}, (ii) isolated single-sound lip-reading \cite{kanamaru2023isolated}, (iii) analyzing facial dynamics derived from speech \cite{Moreira2022}, (iv) and understanding human reactions \cite{Becattini2022}. The following discussion reviews research on the problems of face detection, eye tracking, blinking, and facial expression recognition.

\paragraph{Face Detection}\cite{barua2016} introduces a patch-based model for face detection, highlighting the potential of these sensors for low-power vision applications. \cite{lenz2020} exploited event temporal resolution properties to detect faces using eye blinks (characterized by unique temporal signatures) and applied a probabilistic framework for face localization and tracking in indoor and outdoor environments. Ramesh et al. \cite{ramesh2020} presented an event-based method for learning face representations using kernelised correlation filters within a boosting framework for surveillance applications. In \cite{bissarinova2024fes}, the authors introduced the Faces in Event Streams (FES) dataset, providing 689 minutes of annotated event camera data for event-based face detection. To demonstrate its effectiveness, they trained 12 models to predict bounding box and facial landmark coordinates, achieving real-time detection with an mAP50 score of over 90\%. In \cite{himmi2024msevs}, the authors presented a multispectral event-based vision system for face detection (\textit{MS-EVS}). They introduced three neuromorphic datasets (N-MobiFace, N-YoutubeFaces, and N-SpectralFace), combining RGB videos with simulated color events and multispectral videos. The authors showed that early fusion of multispectral events improved face detection performance over conventional multispectral images, demonstrating the value of multispectral event data. 

\paragraph{Eye Blinking/Tracking}In \cite{lenz2020}, the authors used blinks to track errors in faces through a generic temporal model correlated with local face activity. \cite{angelopoulos2021} proposed a hybrid event-based near-eye gaze tracking system with update rates beyond 10,000 Hz, integrating an online 2D pupil fitting method that updates a parametric model using a polynomial regressor to estimate the gaze point. \cite{wu2025brat} proposed BRAT, an event-based eye-tracking method combining CNN features with Bi-GRU and attention, achieving state-of-the-art accuracy on ThreeET-plus. \cite{huang2025exploring} introduced TDTracker, which uses 3D CNNs and GRU-based modules to capture temporal dynamics and ranked third in the CVPR 2025 challenge. \cite{truong2025dual} presented KnightPupil, integrating EfficientNet-B3, bidirectional GRU, and a state-space module to enhance robustness against abrupt eye movements, with competitive results on 3ET+.

\paragraph{Facial Expression Recognition}
In \cite{barchid2024spikingfer}, the authors presented \textit{Spiking-FER}, a Spiking Neural Network (SNN) for facial expression recognition. The original datasets were downsampled to 200×200 pixels and converted to event sequences using neuromorphic emulation tools. Results show the SNN achieves performance comparable to conventional deep neural networks while reducing energy consumption (up to 65.39 times). The NEFER dataset \cite{berlincioni2023nefer} combines RGB videos and events for emotion recognition, capturing micro-expressions difficult to detect with RGB data but evident in event data. Results show that the event-based approach doubles accuracy compared with RGB-based methods, demonstrating neuromorphic effectiveness in analyzing fast expressions. \cite{verschae2023eventgesturefer} presents an analysis of event-based gesture and facial expression recognition. Deep Learning-based classification models were evaluated across various settings using gesture-recognition databases and two synthetic facial-expression databases (e-MMI and e-CK+).

\section{Preliminaries}
\label{sec:preliminaries}

This section provides an overview of event-based cameras and synthesizes research on data representations, classification methods, neuromorphic datasets, and tools for event-based data generation, including simulators and emulators. This description is required to introduce the proposed method and its evaluation.

\begin{figure}[!h]
    \centering
    \includegraphics[width=1.0\linewidth]{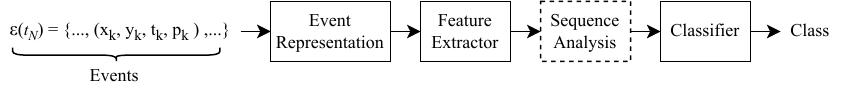}
    \caption{Pipeline for event-based classification that addresses the need of (i) a representation that captures the event's spatial distribution and asynchronous temporal evolution, (ii) a feature extractor that properly encodes the input information, (iii) an analysis over a long sequence of events, (iv) a classifier module.}
    \label{fig:proposal-arch}
\end{figure}

Most studies that have addressed classification problems of event data have used a process similar to that shown in Figure~\ref{fig:proposal-arch}. This process involves an event representation that captures spatial and temporal progression, a feature extractor, a classification layer, and sometimes a sequence module for analyzing temporal patterns.

\subsection{Event generation model}

Event cameras are sensors where each pixel independently responds to changes in the received logarithmic brightness signal $L(\textbf{u}_k, t_k) \doteq \text{log}( I(\textbf{u}_k, t_k) )$, where $\textbf{u}_k$ is the sensor pixel, $t_k$ is the associated time, $I$ is the brightness signal, and $L$ is the log-brightness  \cite{Gallego2019-wf}. An event is triggered at pixel $\textbf{u}_k = (x_k, y_k)^T$ and time $t_k$ when the brightness change since the last event reaches a threshold $\pm C > 0$: 
\begin{equation}
    L(\textbf{u}_k, t_k) - L(\textbf{u}_k, t_k - \Delta t_k) = p_k C, 
\end{equation}
where $p_k \in \{-1, +1\}$ is the brightness change sign, and $\Delta t_k$ is the time since the last event at $\textbf{u}_k$. In a time interval, an event camera produces a sequence $\mathcal{E}(t_N) = \{ e_k \}_{k=1}^{N} = \left\{ (\textbf{u}_k, t_k, p_k)\right\}_{k=1}^{N}$, where $t_N$ denotes the last timestamp in sequence $\mathcal{E}$.

\subsection{Event representations}
\label{related_work:representations}

A critical issue is extracting meaningful information from event sequences. The basic level of representation is an individual event, consisting of coordinates, timing, and polarity. A second level is an event packet, which corresponds to a spatiotemporal neighborhood of events. Some representations accumulate events in the image plane. Other representations are obtained by preprocessing events into dense images or tensors for image-based models, which are then used in convolutional neural networks.

Some prior representations include \cite{Gallego2019-wf}: \begin{inparaenum}[a)]
\item	Event frame or 2D histogram, 
\item	Time surface: 2D map where pixels store a time value,
\item	Voxel Grid: space-time 3D histogram of events, where voxels represent pixels and time intervals,
\item	3D point set: events in the spatio-temporal neighborhood processed as points in 3D space,
\item	Point sets on image plane: Events as evolving 2D points on image plane,
\item	Motion-compensated image: representation based on events and motion hypothesis, and
\item	Reconstructed images: a more motion-invariant representation of event frames.
\end{inparaenum}

These representations do not fully leverage the rich spatiotemporal information of event sequences and hinder fast processing by requiring the integration of information across time. Thus, another approach uses representations that characterize the spatiotemporal distribution of events, such as the Event Spike Tensor \cite{gehrig2019est}, Sparse Recursive Representation \cite{messikommer2020asynet}, Temporal Binary Representation \cite{innocenti2021tbr, baldwin2020inceptive}, Time-Ordered Recent Event Volumes \cite{baldwin2021tore}, Neighborhood Suppression Time Surface \cite{Wan2021-tj}, Binary Event History Image \cite{wang2022behi}, and Tencode \cite{huang2023tencode}. In the following section, we describe SSR and EST, two relevant representations considered later in this article.

\paragraph{Sparse Recursive Representations (SSR)}
In \cite{messikommer2020asynet}, the authors proposed an image-like representation $H_{t_N} (\textbf{u}, c)$ for an event sequence $\mathcal{E}(t_N)$, where $\textbf{u} = (x, y)^T$ denotes a pixel, and $c$ denotes a channel. This representation can be processed by standard CNNs and preserves the spatial sparsity of events, but discards the temporal sparsity.
The authors propose to recover the temporal sparsity by focusing on the change in $H_{t_N} (\textbf{u}, c)$ when a new event arrives:
\begin{equation}
    H_{t_{N+1}} (\textbf{u}, c) = H_{t_{N}} (\textbf{u}, c) + \sum_{i}^{}\Delta_i(c) \delta(\textbf{u} - \textbf{u}^{'}_i),
\end{equation}
leading to increments $\Delta_ i(c)$ at only a few positions $\textbf{u}^{'}_i$ in $H_{t_N} (\textbf{u}, c)$, to maximize efficiency, and using Dirac pulses $\delta(\textbf{u} - \textbf{u}^{'}_i)$.

\paragraph{The Event Spike Tensor (EST)}In \cite{gehrig2019est}, the authors converted the event set $\mathcal{E}(t_N)$ into a grid-based representation, preserving spatial, temporal, and polarity information. To use this in standard deep learning networks (e.g., CNNs), it is necessary to map sequence $\mathcal{E}$ to tensor $\mathcal{T}$, that is, $\mathcal{M}: \mathcal{E} \rightarrow \mathcal{T}$.

Given that events represent point sets in a four-dimensional manifold spanned by the $x$ and y spatial coordinates, time, and polarity, 
\cite{gehrig2019est} proposes a grid representation: 
\begin{equation}
\begin{split}
    S_{\pm} [x_l, y_m, t_n] &  = \sum_{e_k \in  \mathcal{E}_{\pm}}
    f_{\pm}(x_k,y_k,t_k) h(x_l - x_k, y_m - y_k, t_n - t_k),
\end{split}
\label{eq:EST}
\end{equation}
where $f_{\pm}$ is an event Measurement Field (e.g., encoding temporal or polarity information) and $h$ is an aggregation kernel.

Typically, the spatiotemporal coordinates, $x_l, y_m, t_n$, lie on a voxel grid, that is, $x_l \in \{0, 1, ..., W - 1\}, y_m \in \{0, 1, ..., H - 1\}$, and $t_n \in \{t_0, t_0 + T, ..., t_0 + B  T \}$, where $t_0$ is the first timestamp, $T$ is the bin size, and $B$ is the number of temporal bins. 

\subsection{Event-based classification methods}

Several models have been designed to capture the temporal evolution of events, including SNNs, graph-based convolutional networks \cite{bi2019} and asynchronous event processing \cite{messikommer2020asynet}. In this subsection, we examine two learning models associated with the previously reviewed representations \cite{gehrig2019est, messikommer2020asynet} and summarize the other relevant classification models.

For end-to-end representation learning of asynchronous event-based data, in \cite{gehrig2019est} EST feeds a CNN with ResNet-34 as backbone. It uses a fully connected layer to detect processed features and a softmax activation function to assign a probabilistic distribution to each class. This architecture has been used for object recognition (e.g. in the N-Caltech101 dataset \cite{orchard2015convertingdatasets}) and gesture recognition (e.g. the DVS128 Gesture Dataset \cite{amir2017ibm}). In \cite{verschae2023eventgesturefer}, the method achieved 93.82\% accuracy in event-based gesture recognition.

A second method is Asynchronous Sparse Convolutional Networks (Asynet) \cite{messikommer2020asynet}, in which a sparse representation of events and a Sub-manifold Sparse Convolutional (SSC) network are learned. This method exploits spatio-temporal sparsity in classical convolutional architectures by reducing computational power and processing events asynchronously. The authors described two variants: \textit{Dense} and \textit{Sparse}. Both variants process sparse representations asynchronously but differ in that the first uses a traditional CNN, whereas the second uses the SSC network. \cite{verschae2023eventgesturefer} reports that on the DVS128 Gestures dataset, the Asynet architecture outperforms end-to-end representational learning method \cite{gehrig2019est} in gesture recognition, achieving 94.66\% accuracy versus 93.82\%. According to the authors, this is mainly due to the SSR and SSC networks for asynchronous events; however, this method has high inference time.

Studies on event-based cameras have explored approaches for designing novel spatio-temporal learning models. \cite{shrestha2018, kaiser2020, kaiser2020decolle} employed SNNs in classification tasks to exploit the spatio-temporal distribution of events, responding to signals exceeding activation thresholds. \cite{Lagorce2017-ka, Sironi2018-si} used traditional classification models, such as SVMs, and benefited from adaptive kernels that capture temporal data evolution. \cite{Gammulle2017-stream} adopted a CNN-based architecture coupled with an LSTM layer to capture events' spatial and temporal distribution. \cite{2020Pansuriya} incorporated a Deep Recurrent Neural Network for human activity recognition, where the network becomes deep in time and spatial domains.

\subsection{Datasets and data generation using simulators/emulators of event-based cameras}

A key challenge in exploring new supervised learning tasks is the availability of databases. For event-based facial analysis, databases are limited, including: for face detection are Face Detection dataset \cite{lenz2020}, N-MobiFace \cite{himmi2024msevs}, N-YoutubeFaces \cite{himmi2024msevs}, and N-SpectralFace \cite{himmi2024msevs}, while the DAVIS-Face dataset \cite{ramesh2020} and Neuromorphic-Helen dataset \cite{ryan2020realtimeface} are available for eye tracking and blink recognition. For face expression recognition, recent datasets include NEFER \cite{berlincioni2023nefer}, DVS-CK+ \cite{barchid2024spikingfer}, DVS-ADFES \cite{barchid2024spikingfer}, DVS-CASIA \cite{barchid2024spikingfer}, DVS-MMI \cite{barchid2024spikingfer}, e-CK+ \cite{verschae2023eventgesturefer} and e-MMI \cite{verschae2023eventgesturefer}.

Given the lack of data on many problems involving event-based cameras, some datasets have been generated using simulators/emulators, such as N-MobiFace, N-YoutubeFaces, e-CK+, and e-MMI. Two methods for generating realistic events from video frames are ESIM \cite{billard2018esim} and V2E \cite{hu2021v2e}. ESIM simulates an event camera using an adaptive rendering scheme that samples frames as needed \cite{billard2018esim}. The V2E simulator aims to replicate event cameras by generating realistic DVS event sequences with high temporal resolution (using an interpolation model \cite{niklaus2017superslomo}), high dynamic range (by computing the luminance intensity logarithm), pixel-level Gaussian event-threshold mismatch, a finite intensity-dependent bandwidth, and intensity-dependent noise \cite{hu2021v2e}. In \cite{verschae2023eventgesturefer}the e-CK+ dataset was introduced, in which emulated events were generated from frames using V2E, and an event-frame pair was available for each sample.

\section{evTransFER: a framework for event-based facial expression recognition}
\label{sec:methodology}

\begin{figure}[!ht]
    \centering
    \includegraphics[trim={0.4cm 0 0 0},clip,width=\linewidth]{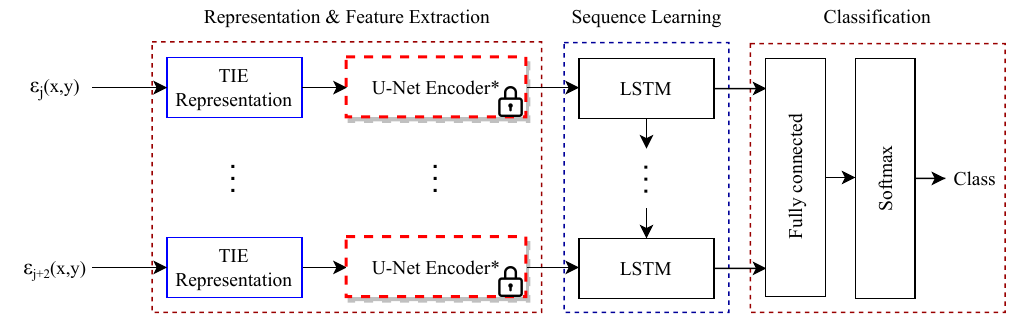}
    \caption{Proposed Architecture for event-based facial expression recognition, comprising (i) a new event representation called TIE (Temporal Information of Events), (ii) a feature extractor module for identifying and encoding events' key patterns (encoder trained for face reconstruction, see Section~\ref{sec:methodology-encoded}), (iii) temporal memory networks enabling temporal learning of event sequences, and (iv) a classification module that categorizes events into facial expressions.
    }
    \label{fig:EST-LSTM-arch}
\end{figure}

The present section describes the proposed learning framework, evTransFER. 
This framework addresses: \begin{inparaenum}[a)]\item the design of a representation that considers the temporal distribution of events, \item the use of an encoder as a feature extractor obtained by transfer learning (to have prior knowledge of the task), and\item the integration of a spatio-temporal sequential module,\end{inparaenum}and it is based on the architecture shown in Figure~\ref{fig:EST-LSTM-arch}. The modules and training methodologies of the proposed architecture are described in the following sections.

\subsection{Proposed Temporal Information of Events (TIE) representation}

We begin by outlining the proposed Temporal Information of Events (TIE) representation, which is grounded in the Event Spike Tensor \cite{gehrig2019est} mentioned earlier.

\paragraph{Temporal encoding} 

As outlined in Section~\ref{related_work:representations}, EST is a representation that captures spatio-temporal information of events within a voxel grid data structure. The analytical expression of the voxel grid corresponds to sampling the convolved signal at regular intervals (see Equation~\ref{eq:EST}). A challenge with this representation stems from temporal coding and lack of timestamp normalization in the event sequence, which affects the measure function $f_{\pm}(x_k,y_k,t_k)$ and kernel $h(x_l - x_k, y_m - y_k, t_n - t_k)$. 
We observe that temporal encoding and normalization methods significantly influence classification accuracy. To address this, we define the temporal variables as 
$\tau_k = \frac{t_k}{t_N}$, 
$\hat{\tau}_k = \frac{t_k - t_1}{t_N - t_1}$
and establish two measure functions based on $\tau_k$ and $\hat{\tau}_k$: 
\begin{equation}
    \label{eq:ftau}
    \begin{aligned}
        f_{\tau_k} &= f_{\pm}(x_k,y_k,\tau_k)  = \tau_k = \frac{t_k}{t_N}, \\
        f_{\hat{\tau}_k} &= f_{\pm}(x_k,y_k,\hat{\tau}_k) = \hat{\tau}_k = \frac{t_k - t_1}{t_N - t_1},
    \end{aligned}
\end{equation}
then the kernel can be evaluated in $\tau_k$ and $\hat{\tau}_k$ (see Equation~\ref{eq:ftau}):
\begin{align}
\label{eq:htau}
h_{\tau_k} &= 
\begin{aligned}[t]
  &h(x_l - x_k, y_m - y_k, \tau_n - \tau_k), \\
\end{aligned} \\
\label{eq:htauhat}
h_{\hat{\tau}_k} &= 
\begin{aligned}[t]
  &h(x_l - x_k, y_m - y_k, \hat{\tau}_n - \hat{\tau}_k), \\
\end{aligned}
\end{align}
with $\tau_n - \tau_k = \frac{t_n - t_k}{t_N}$ and 
$\hat{\tau}_n - \hat{\tau}_k = \frac{t_n - t_k}{t_N - t_1}$.

\paragraph{Tensor mapping}

The EST representation converts event sequences into grid-based structures, effectively preserving spatial, temporal, and polarity information. Due to the dimensional complexity inherent in this structure, we project this representation into images, thereby enabling the utilization of existing deep learning-based architectures designed for three-channel inputs.. To achieve this, we first take the EST and reshape it starting from the tensor $S \in \mathbb{R}^{\left(  C, H, W \right)}$ to a tensor $\hat{S} \in \mathbb{R}^{\left( C, \frac{2 C}{3}, H, W \right)}$:
\begin{equation}
    S=\left( S^{i,j}_k \right)^{1 \leq i \leq W, 1 \leq j \leq H}_{1 \leq k \leq C},\\
    \hat{S}=\left( \hat{S}^{i,j}_{k,l} \right)^{1 \leq i \leq W, 1 \leq j \leq H}_{ 1 \leq k \leq C, 1 \leq l \leq \frac{2 C}{3}},
\end{equation}
with $C$ the number of input channels (e.g. $C=9$), $H, W$ the height and the width of the representation, respectively, $k$ the channel index, $i$ the horizontal index, $j$ the vertical index, and $l$ the bin index. Then, the values of the $\hat{S}$ bin, grouped along the time axis, are then summed up into a tensor $\hat{Z} \in \mathbb{R}^{\left( 3, H, W \right)}$, which is subsequently normalized:
\begin{equation}
    \hat{Z}=\left( \hat{Z}^{i,j}_k \right)^{1 \leq i \leq W, 1 \leq j \leq H}_{1 \leq k \leq 3}, \quad
    \hat{Z}' = \frac{\hat{Z} - \hat{Z}_m}{\hat{Z}_M - \hat{Z}_m},
\end{equation}
where $\hat{Z}_m$ and $\hat{Z}_M$ are the 1\% and 99\% percentiles of $\hat{Z}$. 

\begin{figure}[!h]
    \centering
    \includegraphics[width=\linewidth, trim=90 180 90 50, clip]{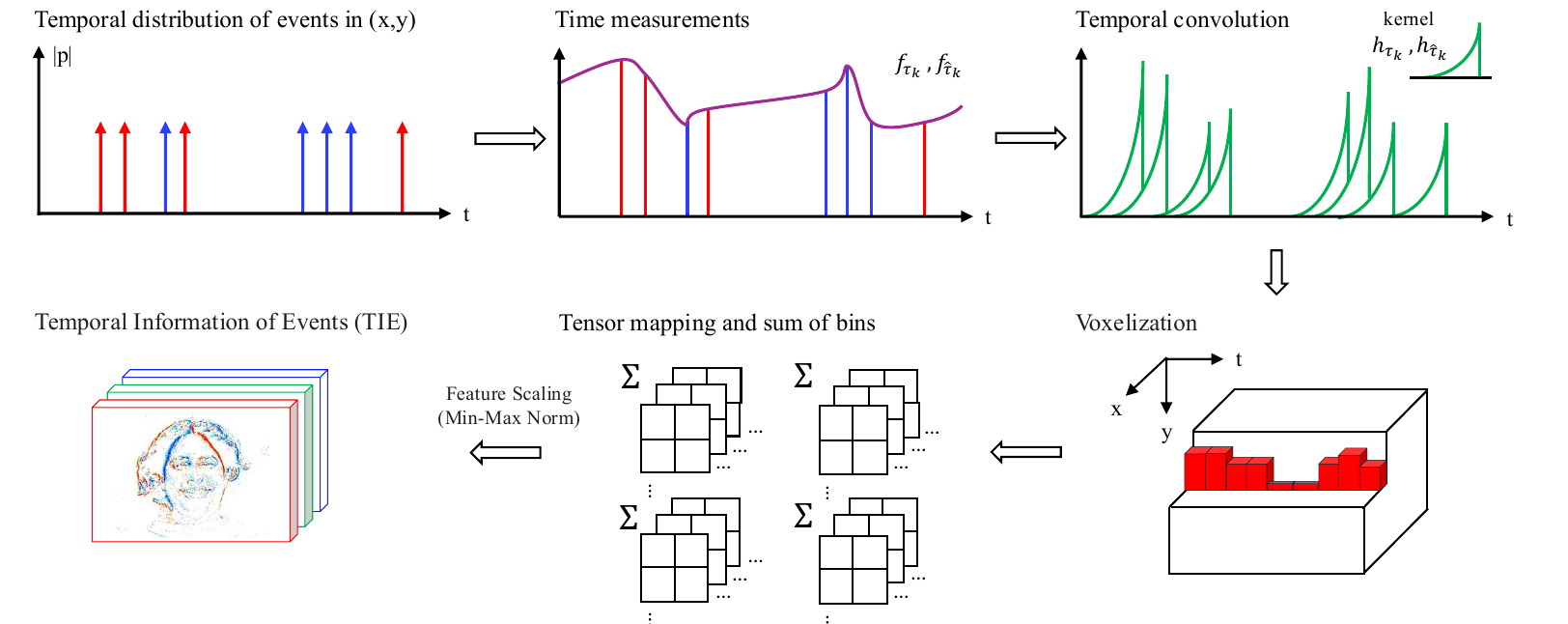}
    \caption{\textit{Pipeline of the proposed representation: Temporal Information of Events} (TIE). TIE is based on the EST representation \cite{gehrig2019est}.}
    \label{fig:tie-representation}
\end{figure}

Finally, $\hat{Z}'$ is mapped from $[0,1] \rightarrow [0,255]$ (saturating values outside the 1\% and 99\% percentiles), to generate the projection in the RGB image space. We refer to our proposed event-based data representation as \textit{Temporal Information of Events} (TIE). Figure~\ref{fig:tie-representation} illustrates the process to obtain the TIE representation.

\begin{figure}[!ht]
\centering
    \subcaptionbox{$f_{\tau_k} \cdot h_{\tau_k}$}
    [.24\textwidth]{\includegraphics[width=\linewidth, trim=0 40 40 0, clip]{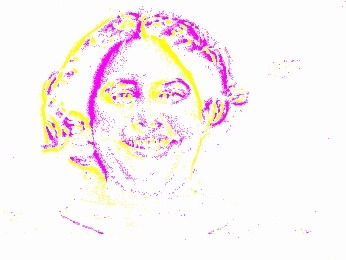}}
    \subcaptionbox{$f_{\tau_k} \cdot h_{\hat{\tau}_k}$}%
    [.24\textwidth]{\includegraphics[width=\linewidth, trim=0 40 40 0, clip]{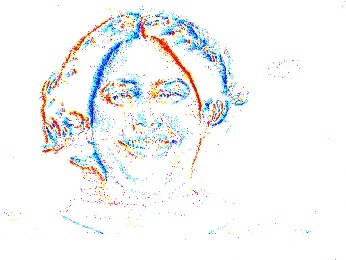}}
    \subcaptionbox{$f_{ \hat{\tau}_k } \cdot h_{\tau_k}$}%
    [.24\textwidth]{\includegraphics[width=\linewidth, trim=0 40 40 0, clip]{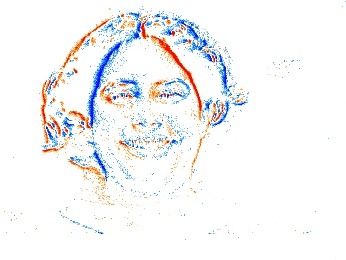}}
    \subcaptionbox{$f_{\hat{\tau}_k } \cdot h_{\hat{\tau}_k }$}
    [.24\textwidth]{\includegraphics[width=\linewidth, trim=0 40 40 0, clip]{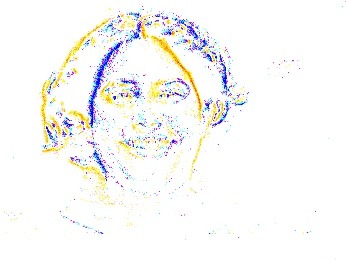}}
    
    \caption{Variants of TIE event representation generated from the e-CK+ dataset by changing the temporal variable $\tau \in \{ \tau_k, \hat{\tau_k}\}$, parameter of measurement function $f$ and the kernel function $h$. 
    }
\label{fig:temporal-and-measurement}
\end{figure}

We recall that ${\tau}$ is a temporal variable that characterizes the distribution of events. In the experiments detailed in Section~\ref{sec:results}, we assess four TIE variants ($f_{\tau_k} \cdot h_{\tau_k}$, $f_{\tau_k} \cdot h_{\hat{\tau}_k}$, $f_{ \hat{\tau}_k } \cdot h_{\tau_k}$, and $f_{\hat{\tau}_k } \cdot h_{\hat{\tau}_k }$), taking into account different normalization factors and kernels. Figure~\ref{fig:temporal-and-measurement} illustrates an example of how these variants are used.

\subsection{Feature extractor training for knowledge transfer}
\label{sec:methodology-encoded}

A crucial element of a feature-rich extractor network is the accurate characterization and encoding of the input information for the learning model. For complex tasks, the network should possess prior knowledge to effectively identify relevant patterns. We propose using event-based facial reconstruction to acquire prior knowledge about face spatio-temporal dynamics, which can then be utilized (as an encoder) in facial expression recognition.
For this approach, we only need samples of event sequences and their corresponding image frames, without requiring additional annotations. This methodology can be readily applied to other problems. This event-based pipeline can be trained using asynchronous events and synchronous frames from a DAVIS346 camera, simulated events from videos, or co-axial imaging setups (which include a beam splitter, an event camera, and an RGB camera, such as the one used in \cite{Liu2025pami}). Then the obtained encoder can provide prior knowledge.

\paragraph{Learning from prior knowledge}

\begin{figure}[!ht]
    \centering
    \includegraphics[width=0.99\linewidth]{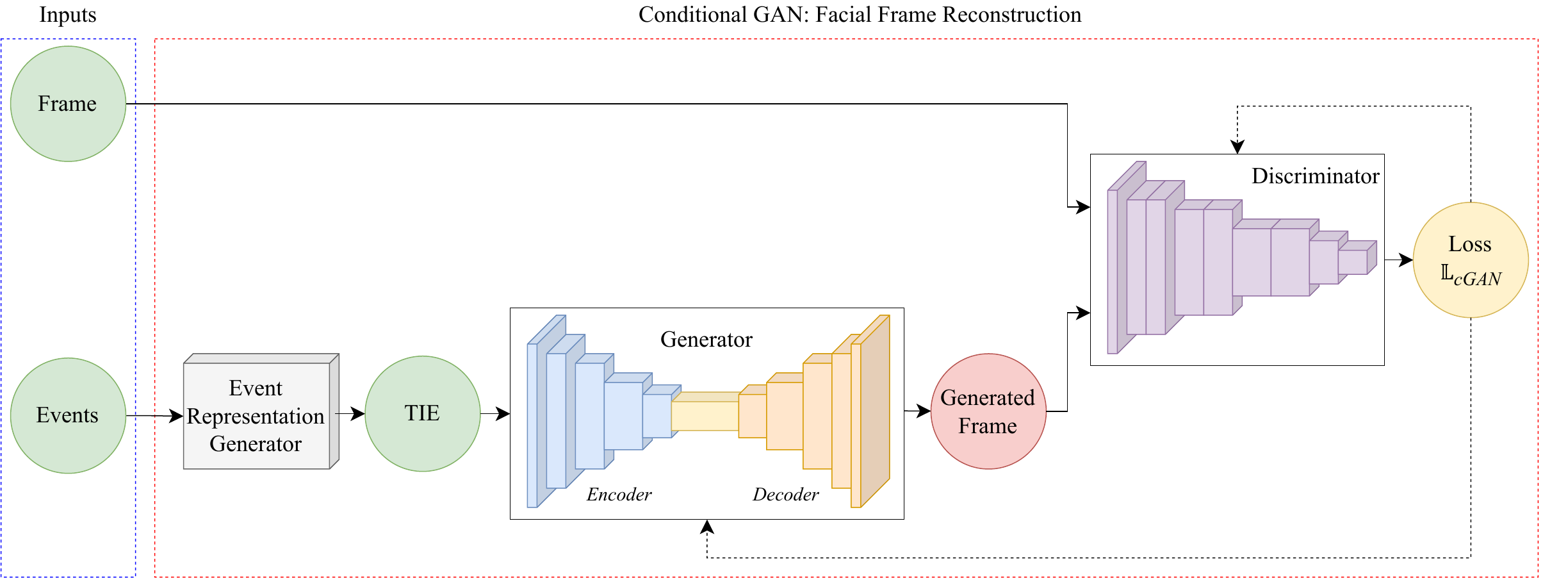}
    \caption{Proposed architecture for event-based facial frame reconstruction.  
    The architecture involves a conditional Generative Adversarial Network that reconstructs frames from a latent variable. This reconstruction system comprises three components: (i) the TIE representation, which serves as a latent variable; (ii) the generator network, which is responsible for reconstructing facial frames from events; and (iii) the discriminator network, which assesses the likelihood of the generated frame against the original frame, providing feedback to the generator based on the loss incurred. The system accepts two inputs: an event sequence and the corresponding frames.
    }
    \label{fig:arch-cGAN}
\end{figure}

The process of reconstructing image frames involves acquiring representations that capture the structural and dynamic characteristics. For facial expressions, this process involves developing a feature extractor by reconstructing facial images. This approach assumes that spatiotemporal dynamics in reconstructed faces are valuable for accurate expression classification.
Our proposal follows these ideas and incorporates the \textit{pix2pix} \cite{isola2016pix2pix} generator encoder for feature extraction. We employ a U-Net-based encoder-decoder \cite{ronneberger2015unet} trained using a conditional Generative Adversarial Network (cGAN) for image reconstruction tasks. This framework uses the TIE representation as a latent variable, with the original frame as the expected output. The cGAN architecture has two networks: a generator that takes a latent variable to replicate a distribution, and a discriminator that distinguishes real samples from generated ones.
The architecture in Figure~\ref{fig:arch-cGAN} serves as an event-based reconstruction network. Subsequently, we use the encoder as a feature extractor for the facial expression classifier.

\begin{figure}[!ht]
	\centering
	\includegraphics[angle=90,width=0.85\linewidth, trim ={48 0 0 0}, clip]{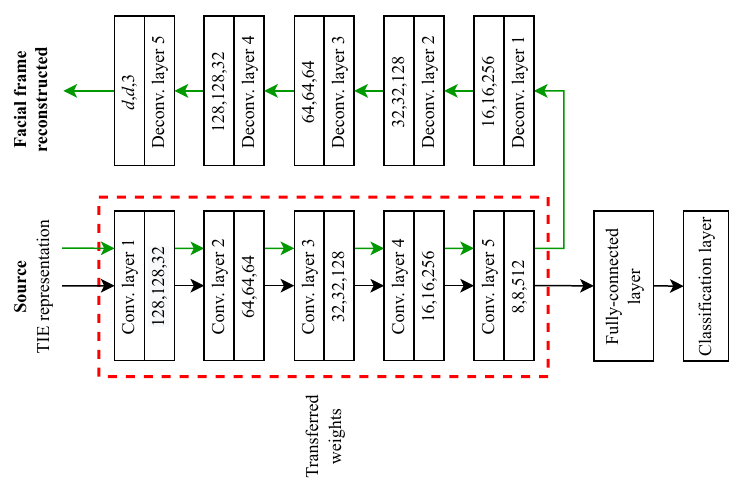}
	\caption{Architecture for training CNN feature extractor in event-based facial frame reconstruction, where prior knowledge and learning transfer for event-based facial expression recognition. The obtained encoder (transferred weights) is shown in red dashed lines, and the reconstruction cycle in green lines. $d,d$ corresponds to the spatial dimension of the reconstructed frame. 
 }
	\label{fig:CNN-arch}
\end{figure}

In the cGAN, the generator $G$ seeks to minimize the difference in the reconstruction of facial expressions from events, while the discriminator $D$ seeks to maximize it. The optimization problem to be solved is 
$    
\min_{G} \max_{D} \mathbb{L}_{cGAN} (G,D), 
$ 
with: 
\begin{equation}
\begin{split}
\mathbb{L}_{cGAN}(G,D) =\;
& \mathbb{E}_{f_{real},f_{gen}}\!\left[\log\!\left(D(f_{real},f_{gen})\right)\right] \\
&+ \mathbb{E}_{e_{TIE}}\!\left[\log\!\left(1-D\!\left(f_{real},G(e_{TIE})\right)\right)\right]
\end{split}
\end{equation}
where $\mathbb{L}_{cGAN}$ is the loss function, $\mathbb{E}$ the expected value, $f_{real}$ the images frames, $e_{TIE}$ the TIE event-based data representation as latent variable, and $f_{gen}$ the generated face frame from the generator autoencoder. By addressing the min-max loss problem, as shown in Figure~\ref{fig:CNN-arch}, we obtain an encoder that uses the TIE representation to reconstruct a facial frame. Once trained for face reconstruction, the encoder is integrated into the proposed facial expression recognition system, as explained in the next section.

\subsection{Proposed Architecture}
\label{sec:methodology-LSTM}

Now we describe in detail the proposed architecture, already shown in Figure~\ref{fig:EST-LSTM-arch}. 
The proposed system receives an $\mathcal{E}_j$ event sub-sequence as input. Following the Stacking Based on Time approach by \cite{rebecq2019highspeed}, the events in-between the time references of two consecutive intensity images of the event camera (denoted as $\Delta t$) are sequenced in $\mathcal{E}_j(x,y)$. Here, the time duration of the event sequence is divided into $n$ equal-scale portions, and then $\mathcal{E}_j(x,y)$ is built by sequencing the events in each time interval $\left[ \frac{(j-1) \Delta t}{n}, \frac{j\Delta t}{n}\right]$. Thus, each $\mathcal{E}_j(x,y)$ sub-sequence is generated from time windows, that is, $\mathcal{E}_j(x,y) = f(j, \Delta t)$, where $f$ is a function that converts event sequences into samples with fixed time windows.
Subsequently, the TIE representation is produced for each event sub-sequence, encoding its spatiotemporal distribution. This representation feeds the U-Net Encoder with static weights (acting as a feature extractor) to learn the temporal sequence of events using an LSTM.
Finally, the classification module uses a Sequence Learning module (LSTMs) connected via a fully connected layer to a softmax to characterize and correlate the sequence of events, thereby classifying the sequence into their respective facial expressions.

With this, we implement a transfer learning from face frame reconstruction to the U-Net encoder. The proposed training procedure, in addition to transferring the trained encoder weights (from the U-Net architecture), fine-tunes all network weights.

In summary, the proposed architecture comprises three key components: the event representation module, feature extractor module, and sequence learning module, implemented using the previously described TIE representation, U-Net-based Encoder, and three LSTM units, respectively. 

\section{Experimental Setup}
\label{sec:evaluationSetUp}

The current section details the experimental setup for training and evaluating the proposed framework, encompassing 
\begin{inparaenum}[a)]
\item the two datasets utilized,
\item the training parameters for encoder learning, and
\item the training parameters for the evaluated state-of-the-art methods.
\end{inparaenum}
In Section~\ref{sec:results}, we compare our approach with state-of-the-art methods and conduct an ablation study on it. Section~\ref{sec:discussion} discusses the results obtained.

\subsection{Datasets}
\label{sec:datasets}

\subsubsection{The Event CK+ dataset}

The \textit{Cohn-Kanade Extended (CK+)} dataset \cite{lucey2010ck+} is among the most frequently used frame-based benchmark datasets for facial expression recognition. Initially, the Cohn-Kanade dataset \cite{kanade2000ck} was created for Action Unit and Valence-Arousal recognition; however, the authors subsequently expanded it to include facial expression analysis.

\begin{figure}[!h]
\centering
    \subcaptionbox{}%
    [0.24\textwidth]{
    \includegraphics[width=\linewidth, trim=0 50 0 15, clip]{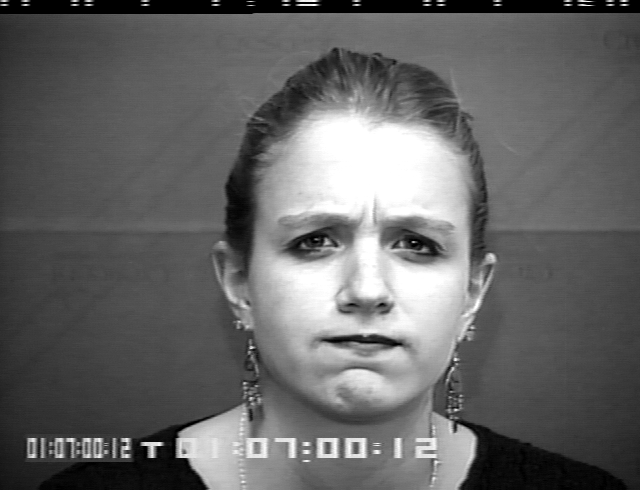}\\
    \includegraphics[width=\linewidth, trim=0 15 0 5, clip]{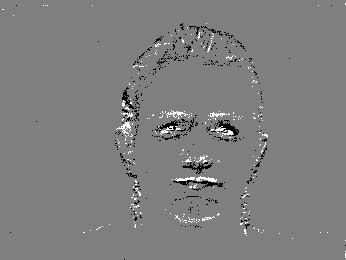}}
    \subcaptionbox{}%
    [0.24\textwidth]{
    \includegraphics[width=\linewidth, trim=0 50 0 15, clip]{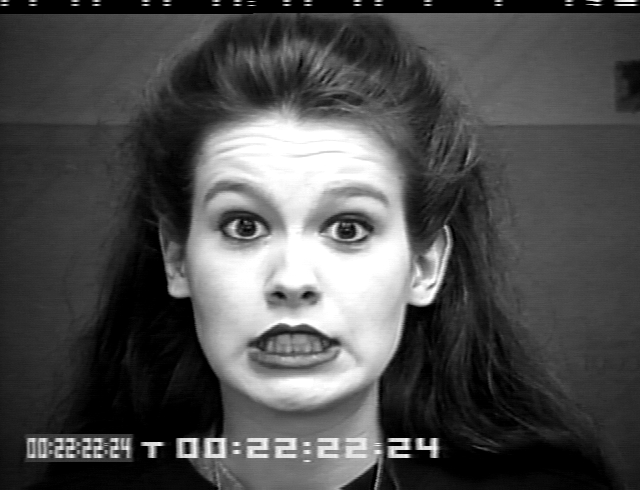}\\\includegraphics[width=\linewidth, trim=0 15 0 5, clip]{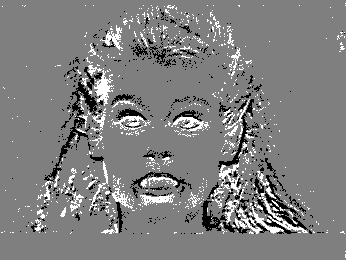}}
    \subcaptionbox{}%
    [0.24\textwidth]{\includegraphics[width=\linewidth, trim=0 50 0 15, clip]{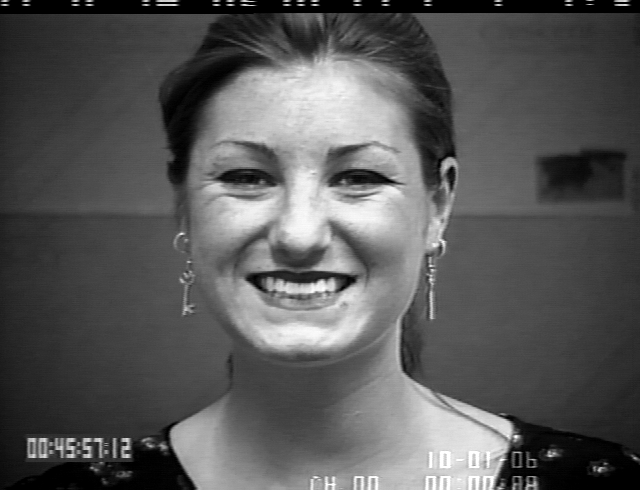}\\
    \includegraphics[width=\linewidth, trim=0 15 0 5, clip]{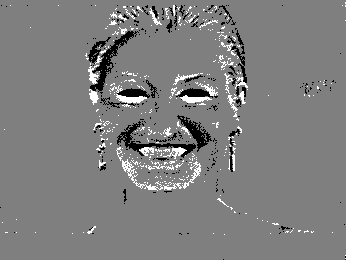}}
    \subcaptionbox{}
    [0.24\textwidth]{\includegraphics[width=\linewidth, trim=0 50 0 15, clip]{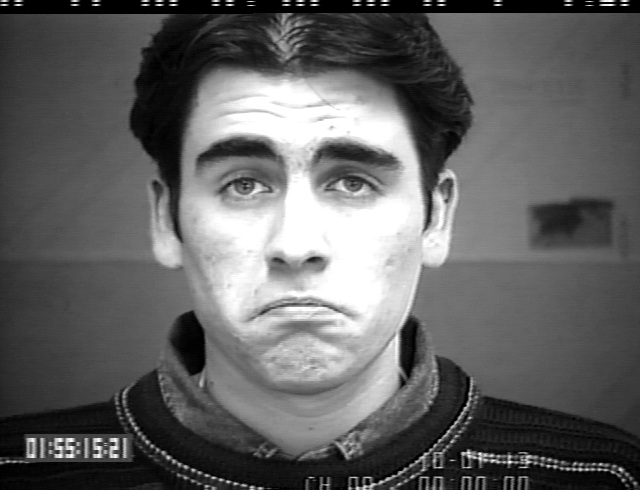}\\\includegraphics[width=\linewidth, trim=0 15 0 5, clip]{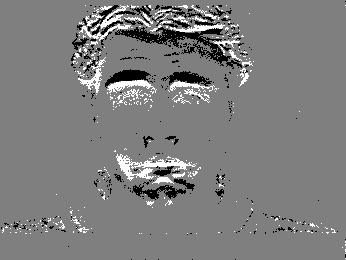}}

\caption{The Event-based Extended Cohn-Kanade Dataset (e-CK+) \cite{verschae2023eventgesturefer}: A event-based dataset for individual facial expression recognition. Examples from the e-CK+ dataset\cite{kanade2000ck, lucey2010ck+}: (a) Anger, (b) Fear, (c) Happiness, (d) Sadness. Top: original image. Bottom: Accumulated event representation visualization, where white and black pixels correspond to events with positive and negative polarity, respectively.}
\label{fig:db-CK+}
\end{figure}

CK+ comprises 593 video sequences featuring 123 subjects aged 18-50, representing diverse genders and heritages. Each video corresponds to a facial transition from a neutral expression to a specific peak expression, recorded at 30 frames per second (FPS) with a resolution of either 640x490 or 640x480 pixels. Among these videos, 327 were categorized under one of the following seven expressions: anger, contempt, disgust, fear, happiness, sadness, and surprise ( examples are provided in Figure~\ref{fig:db-CK+}).

Due to the frame-based nature of the image sequences of CK+, we follow \cite{verschae2023eventgesturefer} and use the e-CK+ dataset, an event-based equivalent of CK+ generated via the V2E emulation method \cite{hu2021v2e}, which produces event sequences with high temporal resolution and dynamic range. Table~\ref{tab:v2e-e-CK+} shows the parameters of V2E used to elaborate the database. We build the e-CK+ for a 346x260 (frames and events), emulating the sensor resolution of the DAVIS346 camera.
Before emulating the events, data augmentation of the CK+ database samples was performed, including horizontal flipping and four random rotations (ranging from $-10\degree$ to $10\degree$). Additionally, the frames in this dataset are not landmark-standardized.
\begin{table}[!h]
\centering
\footnotesize
\caption{Parameters used to generate the e-CK+ database \cite{verschae2023eventgesturefer}. This dataset is used to train and validate the event-based facial expression recognition methods.}
\footnotesize
\begin{tabular}{|l|c|}
\hline
Input data                & Frames with 640x480 pixels \\ \hline
DVS timestamp resolution  & 1 ms      \\ \hline
DVS exposure duration     & 5 ms      \\ \hline
Positive/Negative event threshold  & 0.15      \\ \hline
Sigma threshold variation & 0.03      \\ \hline
Emulated event camera device      & DAVIS346 \\ \hline
Output Spatial dimensions                & 346x260   \\ \hline
Cut-off frequency         & 30 Hz     \\ \hline
Output data                & Frames/events \\ \hline
\end{tabular}
\label{tab:v2e-e-CK+}
\end{table}

Therefore, after the event sequence for the 327 CK+ videos was synthetically generated, the class samples were split into training and validation datasets in an 80\%-20\% ratio. Now, following the procedure described in Section~\ref{sec:methodology-LSTM}, a time window $\Delta t = 30 \text{fps}$ $(\sim33ms)$ is used to generate the e-CK+ samples for this work (based on the CK+ fps) and match the events sub-samples with the respective original frames. Example results 
are shown in Figure~\ref{fig:db-CK+}, bottom row. 
Finally, the frames and event representations were spatially cropped at a 1:1 ratio for frame reconstruction and facial expression classification.

\subsubsection{The NEFER dataset}

NEFER \cite{berlincioni2023nefer} is a real-world event-based facial expression database offering three synchronized modalities for each sample: (i) RGB frames captured with a GoPro camera at 60\,fps, (ii) event-derived frame representations calculated with a temporal binning of 15\,ms, and (iii) the original raw event sequence in \texttt{.raw} format recorded using a Prophesee sensor. These modalities enable direct temporal correspondence between frame-based and event-based signals.

This dataset focuses on facial expression and emotion analysis and includes sequences of both spontaneous and posed expressions under varying head poses and illumination conditions. Crucially, the dataset includes only expression annotations and lacks both bounding boxes and facial landmarks. To address this issue, we cropped the face and event regions using the following procedure: to achieve consistent spatial localization across both modalities, we employed a two-stage process: first, we detected faces in RGB frames using a face detector; subsequently, we approximated facial locations in the event domain through DBSCAN-based clustering on event coordinates, followed by manual refinement. With these aligned detections, we estimated the geometric transformation between frames and events and utilized this mapping to generate bounding boxes for the remaining RGB frames, event TIE frames, and raw event windows.

One practical challenge of NEFER is that many samples exhibit very low event activity, such as near-static facial motion. Consequently, we excluded frame-event pairs whose event windows failed to meet a minimal activity threshold, ensuring that the evaluation of evTransFER was based on samples with significant spatiotemporal content. Due to data protection restrictions, we do not reproduce visual examples from NEFER; readers are directed to \cite{berlincioni2023nefer} for illustrations and further details about the dataset.

\subsection{Learned encoder from face image reconstruction}

As described in Section~\ref{sec:methodology-encoded}, we trained the face reconstruction system and extracted its encoder using paired frame–event data from both the e-CK+ and NEFER datasets. For e-CK+, the reconstruction network was trained on the emulated event sequence and the corresponding CK+ frames. For NEFER, raw event sequences were aligned with the GoPro RGB frames provided by the dataset (only the detected face regions were used). This training setup allows the encoder to learn spatiotemporal facial dynamics from both synthetic and real neuromorphic data.

Table~\ref{tab:cgan2} summarizes the architecture and training parameters of the event-based facial frame reconstruction model. The system is trained for 100~epochs using the Adam optimizer with a learning rate of 0.0002 and a binary cross-entropy loss.

\begin{table}[!h]
\caption{Architectures and training parameters of the event-based facial frame reconstruction method.}
\footnotesize
\label{tab:cgan2}
\centering
\begin{tabular}{|l||c|c|}
\hline
\textbf{Parameter}                        & \multicolumn{1}{c|}{\textbf{Generator}} & \multicolumn{1}{c|}{\textbf{Discriminator}} \\ \hline \hline
Architecture                              & U-Net                                            & PatchGAN                                             \\ \hline
Weight Initialization                     & $\mathcal{N}(\overrightarrow{w}|\mu=0, \sigma^2={0.02}^2)$    &  $\mathcal{N}(\overrightarrow{w}|\mu=0, \sigma^2={0.02}^2)$      \\ \hline
Batch Normalization                       & Yes                                              & Yes                                                  \\ \hline
\multirow{2}{*}{Dropout}                                  & Encoder: No                                      & \multirow{2}{*}{No}                                  \\ \cline{2-2}
                                          & Decoder: Yes, 0.5 dropout                        &                                                      \\ \hline
\multirow{2}{*}{Activation Function (Interm. Layers)} & Encoder: LeakyReLU ($\alpha$=0.2)                       & \multirow{2}{*}{LeakyReLU ($\alpha$=0.2)}                   \\ \cline{2-2}
                                          & Decoder: ReLU                                    &                                                      \\ \hline
Activation Function (Last Layer)          & Tanh                                             & Sigmoid                                              \\ \hline
\end{tabular}
\end{table}

\subsection{Training of the event-based facial expression recognition methods}

Table~\ref{tab:methods_params} summarizes the hyperparameters employed in training of evTransFER, as well as the ones of the state-of-the-art methods used in the evaluations in Section~\ref{sec:results}.

\begin{table}[!h]
\centering
\footnotesize
\caption{Training parameters used for event-based facial expression recognition methods. evTransFER: U-Net trained for Face Reconstruction and later Fine-Tuned.}
\label{tab:methods_params}
\begin{tabular}{|l||c|c|c|c|c|}
\hline
\textbf{Parameter}      & \textbf{EST + CNN} \cite{gehrig2019est} & \textbf{Asynet CNN / SSC} \cite{messikommer2020asynet} & \textbf{ViT} \cite{dosovitskiy2020image} &
\textbf{evTransFER*} \\ \hline \hline
Backbone                & ResNet-34      & VGG-16         & VIT B16 & U-Net Encoder \\ \hline 
Epochs / Batch Size                 & 30 / 4             & 30 /32        &  20/4  & 30 /32           \\ \hline
Learning Rate (Adam)    & \multicolumn{4}{c|}{0.0001} \\ \hline
Learning Rate Sched. & ExponentialLR  & ExponentialLR  & None & ExponentialLR \\ \hline
$\gamma$ (LR Scheduler) & 0.5            & 0.1            & None & 0.1 \\ \hline
Loss Function           & \multicolumn{4}{c|}{Cross Entropy} \\ \hline
\end{tabular}
\end{table}

\section{Evaluation}
\label{sec:results}

In the present Section, a comparative evaluation of the event-based facial expression recognition is performed, including:
\begin{inparaenum}[i)]
\item an evaluation of the performance effects of TIE variants and the use of LSTM (Section~\ref{sec:eval}),
\item a comparative analysis of the proposed method and state-of-the-art methods (Section~\ref{sec:sotacom}), and 
\item an ablation study that shows the contribution of the different techniques used in the proposed classification architecture is performed (Section~\ref{sec:ablation}).
\end{inparaenum}

\subsection{TIE representation and Event-based facial frame reconstruction}

To assess the quality of the learned encoder prior to its transfer to the facial expression recognition task, we evaluate the event-based facial frame reconstruction performance on both the e-CK+ and NEFER datasets. 

\begin{figure}[!ht]
\centering
    \includegraphics[width=0.16\linewidth]{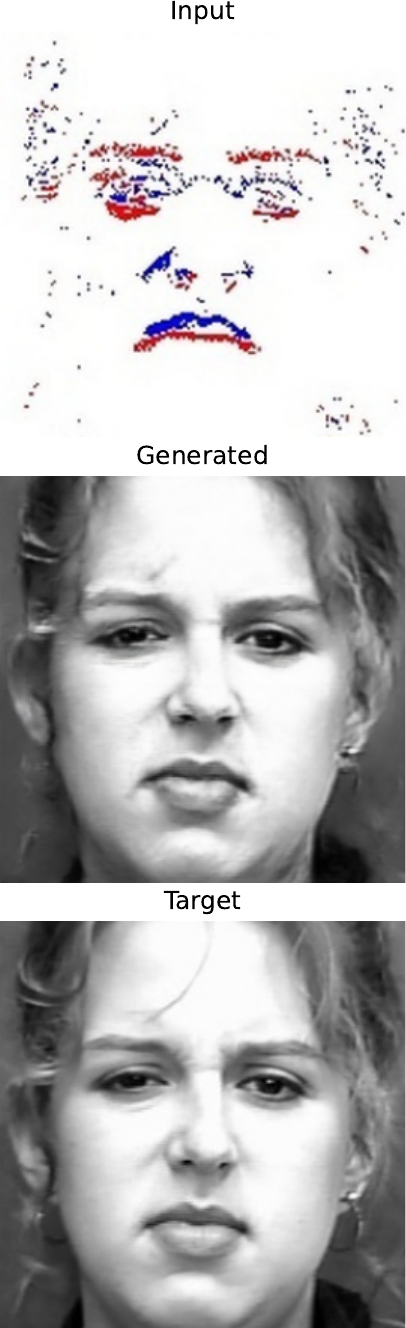}
    \quad
    \includegraphics[width=0.16\linewidth]{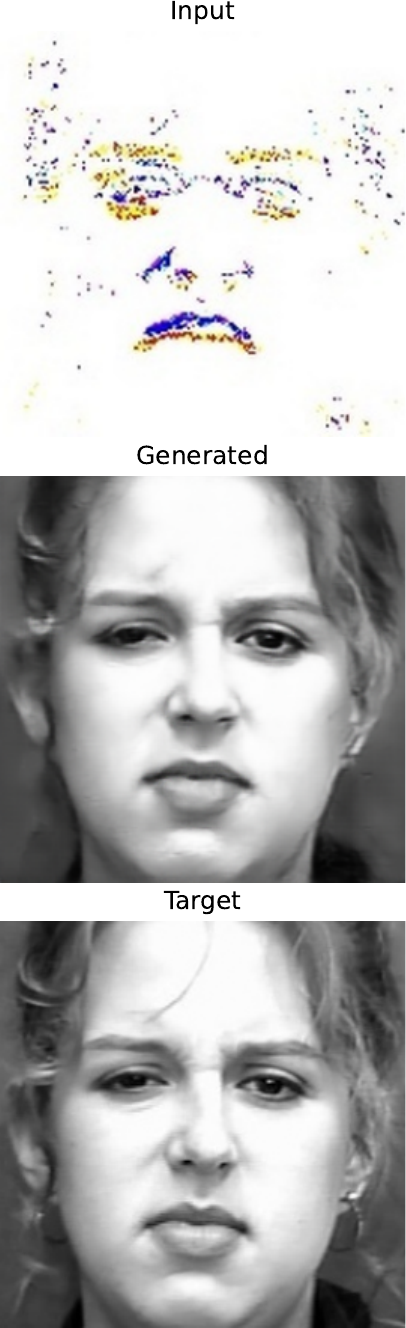}
    \quad
    \includegraphics[width=0.16\linewidth]{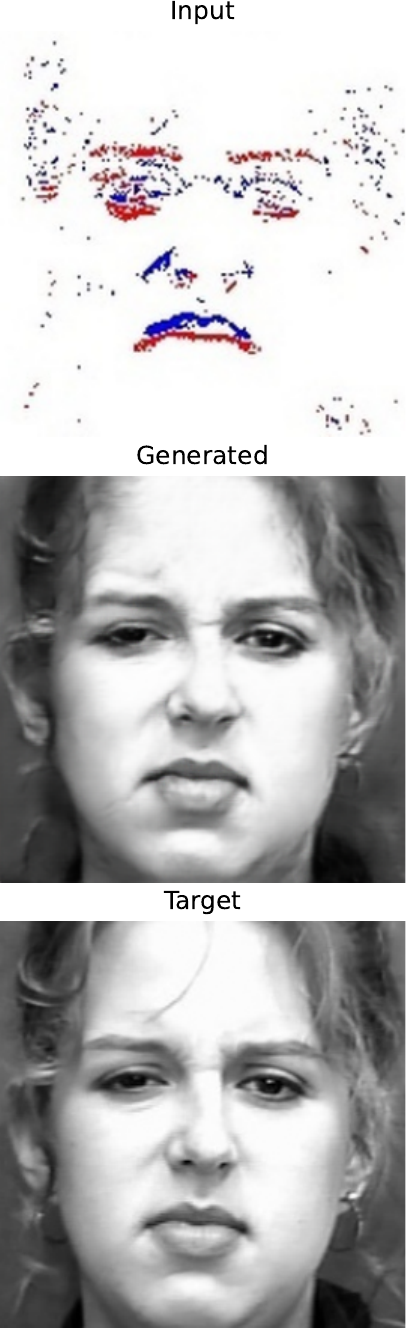}
    \quad
    \includegraphics[width=0.16\linewidth]{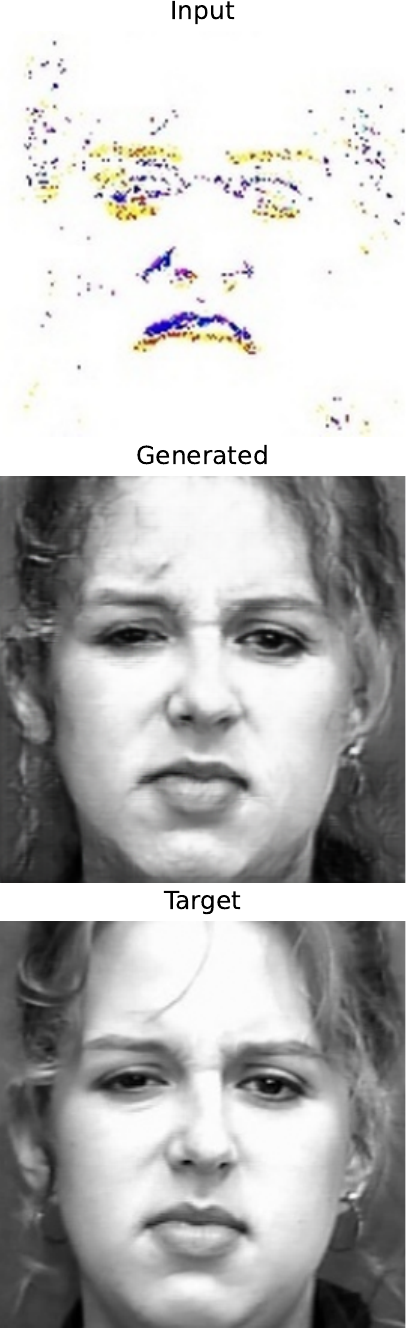}

    \vspace{1em}

    \includegraphics[width=0.16\linewidth]{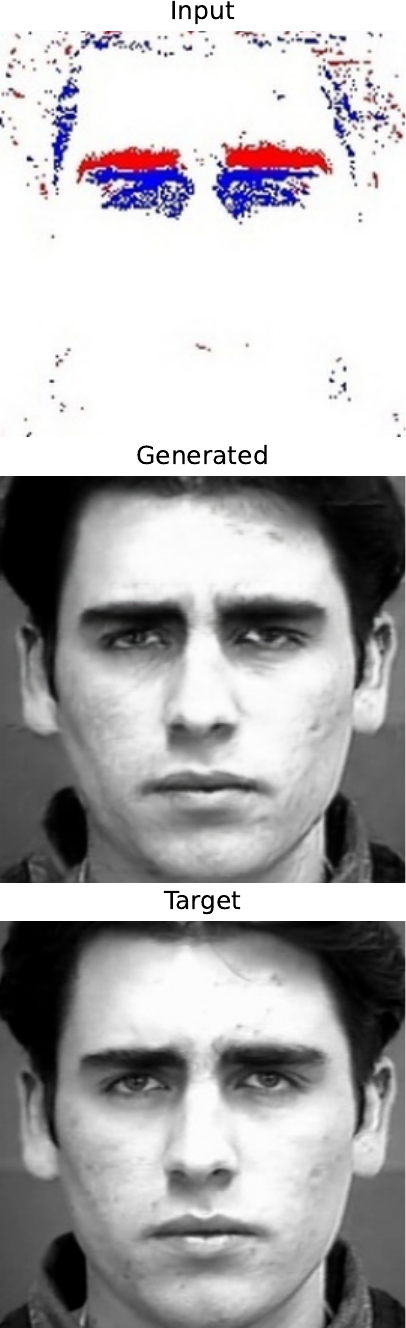}
    \quad
    \includegraphics[width=0.16\linewidth]{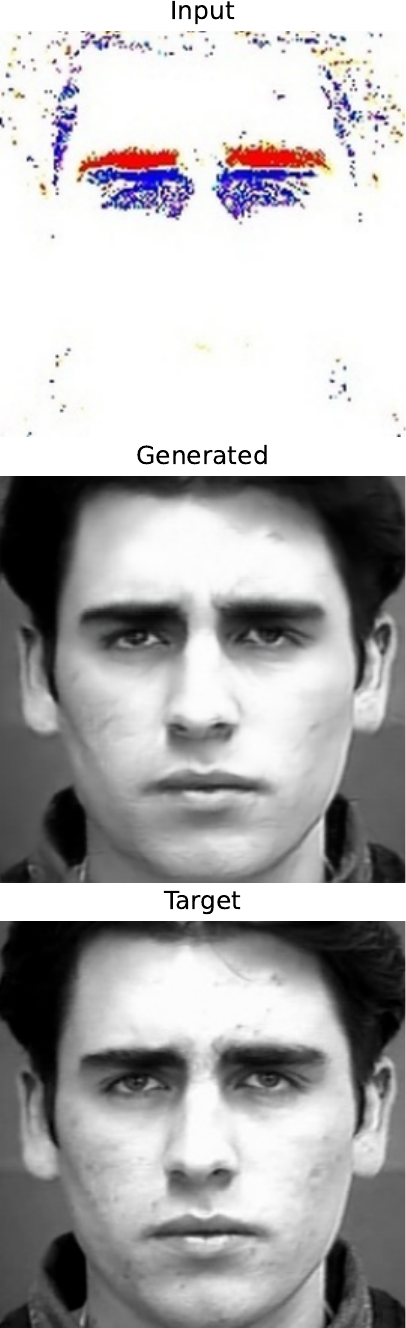}
    \quad
    \includegraphics[width=0.16\linewidth]{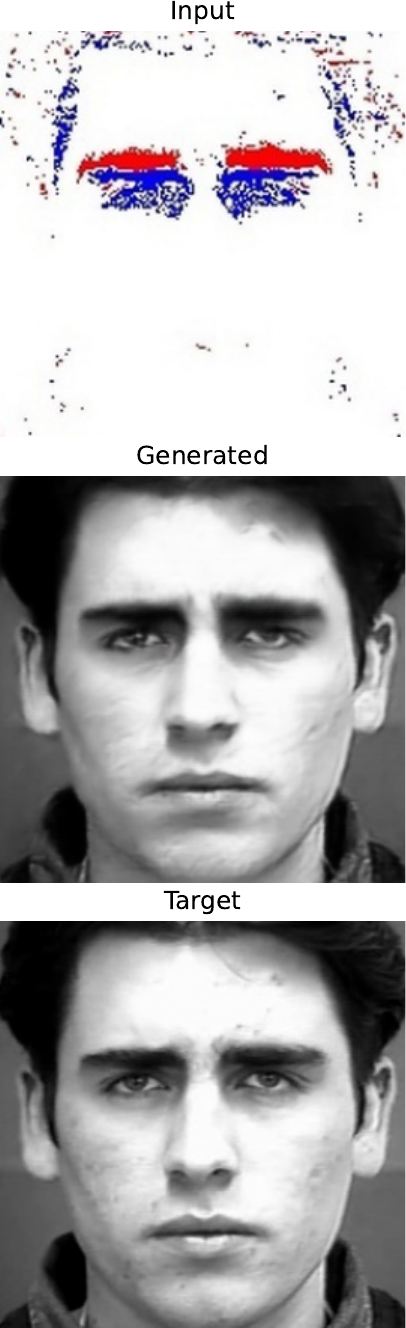}
    \quad
    \includegraphics[width=0.16\linewidth]{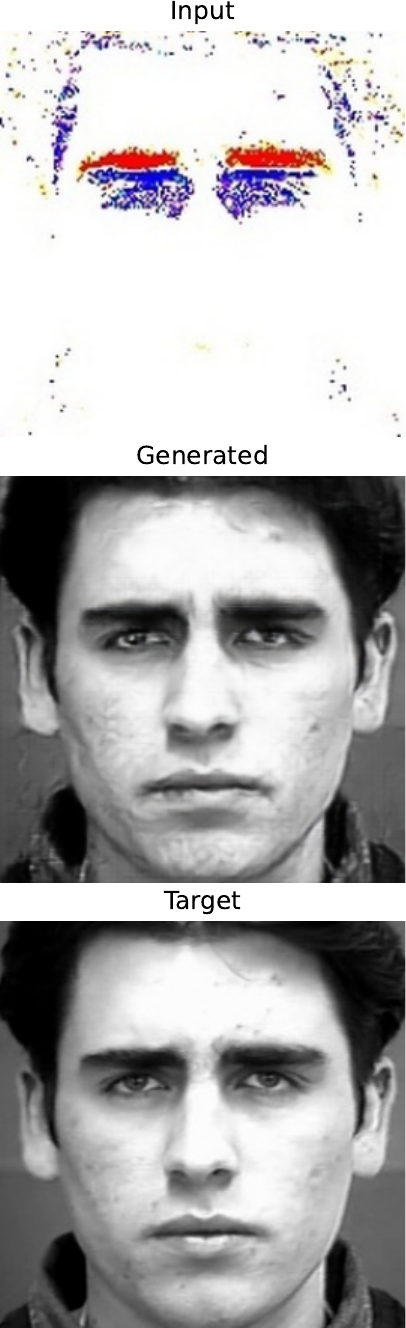}
    
\caption{Event-based facial frame reconstruction results after 80 epochs using the conditional generative adversarial network. Each block shows (top) the event-based input representation, (middle) the generated reconstruction, and (bottom) the target frame. The examples illustrate the model's ability to reconstruct facial structure and expression cues from sparse event patterns.}
\label{fig:results-face-reconstruction}
\end{figure}

The reconstruction network is trained using paired event representations and RGB frames, and the encoder extracted from this model is later reused as the feature extractor in evTransFER. Table~\ref{tab:reconstruction-eck} and Table~\ref{tab:reconstruction-nefer} report quantitative results at epoch~80 for the four temporal-domain and kernel combinations described in Section~\ref{sec:methodology}. Figure~\ref{fig:results-face-reconstruction} provides qualitative examples of the e-CK+ generated frames.

\begin{table}[!h]
\centering
\footnotesize
\caption{Event-based facial reconstruction performance on the \textbf{e-CK+} dataset at epoch 80, comparing temporal-domain and kernel-based combinations. The best values for each metric are shown in bold.}
\begin{tabular}{|l||c|c|c|c|}
\hline
\multirow{2}{*}{Metric} &
\multicolumn{4}{c|}{e-CK+ Facial Reconstruction Results} \\ \cline{2-5}
& \multicolumn{1}{c|}{$f_{\tau_k} \cdot h_{\tau_k}$}    
& \multicolumn{1}{c|}{$f_{\tau_k} \cdot h_{\hat{\tau}_k}$} 
& \multicolumn{1}{c|}{$f_{\hat{\tau}_k} \cdot h_{\tau_k}$}
& $f_{\hat{\tau}_k} \cdot h_{\hat{\tau}_k}$ \\ \hline \hline

MSE  & 0.0017 & \textbf{0.0010} & 0.0015 & 0.0020 \\ \hline
RMSE & 0.0403 & \textbf{0.0319} & 0.0388 & 0.0446 \\ \hline
MAE  & 0.0271 & \textbf{0.0210} & 0.0266 & 0.0301 \\ \hline
PSNR & 27.9830 & \textbf{30.0284} & 28.3061 & 27.1474 \\ \hline
SSIM & 0.8477 & \textbf{0.8826} & 0.8640 & 0.8188 \\ \hline
NCC  & 0.9897 & \textbf{0.9927} & 0.9903 & 0.9869 \\ \hline

\end{tabular}
\label{tab:reconstruction-eck}
\end{table}

As shown in Table~\ref{tab:reconstruction-eck}, the $f_{\tau_k} \cdot h_{\hat{\tau}_k}$ representation consistently yields the best e-CK+ facial reconstruction quality across all metrics (MSE, RMSE, MAE, PSNR, SSIM, and NCC). This indicates that normalizing timestamps with respect to the global event window ($\hat{\tau}_k$) while filtering with a kernel defined in the absolute temporal domain ($h_{\tau_k}$) provides the most stable temporal encoding for emulated event data. Qualitative results in Figure~\ref{fig:results-face-reconstruction} confirm that this representation preserves facial structure more accurately and reduces temporal noise introduced by rapidly changing event rates.

A similar performance is observed for NEFER (Table~\ref{tab:reconstruction-nefer}), despite the increased difficulty of reconstructing from raw, real event sequence. Again, $f_{\tau_k} \cdot h_{\hat{\tau}_k}$ achieves the strongest performance across all metrics, with clear improvements over other variants. Real event data naturally exhibit greater sparsity, device noise, and nonlinear motion dynamics, which explains lower absolute PSNR and SSIM values than e-CK+. Nevertheless, the results demonstrate that the temporal normalization used in $h_{\hat{\tau}_k}$ effectively compensates for non-uniform event rates and exposure differences present in real neuromorphic samples.

\begin{table}[!h]
\centering
\footnotesize
\caption{Event-based facial reconstruction performance on the \textbf{NEFER} dataset at epoch 80, comparing temporal-domain and kernel-based combinations. The best values for each metric are shown in bold.}
\begin{tabular}{|l||c|c|c|c|}
\hline
\multirow{2}{*}{Metric} &
\multicolumn{4}{c|}{NEFER Facial Reconstruction Results} \\ \cline{2-5}
& \multicolumn{1}{c|}{$f_{\tau_k} \cdot h_{\tau_k}$}    
& \multicolumn{1}{c|}{$f_{\tau_k} \cdot h_{\hat{\tau}_k}$} 
& \multicolumn{1}{c|}{$f_{\hat{\tau}_k} \cdot h_{\tau_k}$}
& $f_{\hat{\tau}_k} \cdot h_{\hat{\tau}_k}$ \\ \hline \hline

MSE  & 0.0030 & \textbf{0.0019} & 0.0025 & 0.0055 \\ \hline
RMSE & 0.0541 & \textbf{0.0440} & 0.0495 & 0.0715 \\ \hline
MAE  & 0.0389 & \textbf{0.0297} & 0.0326 & 0.0532 \\ \hline
PSNR & 25.4543 & \textbf{27.2208} & 26.2032 & 23.2468 \\ \hline
SSIM & 0.8127 & \textbf{0.8358} & 0.8226 & 0.7564 \\ \hline
NCC  & 0.9851 & \textbf{0.9867} & 0.9833 & 0.9755 \\ \hline

\end{tabular}
\label{tab:reconstruction-nefer}
\end{table}

\subsection{Proposed face expression recognition method evaluation}
\label{sec:eval}

We now assess the proposed framework, evTransFER, using the CK+ and NEFER datasets. Tables~\ref{tab:TIE-performance-0} and~\ref{tab:TIE-performance-NEFER} present the consolidated results for the four TIE variants, both with and without LSTM in evTransFER. Both tables indicate that LSTM generally performs best for most variants in the CK+ dataset. Table~\ref{tab:TIE-performance-0} further reveals that the $f_{\hat{\tau}_k} \cdot h_{\tau_k}$ variant is the most effective for CK+, whereas for NEFER, the variants $f_{\tau_k} \cdot h_{\hat{\tau}_k}$ and $f_{\hat{\tau}_k} \cdot h_{\hat{\tau}_k}$ yield very similar results. Given that $f_{\hat{\tau}_k} \cdot h_{\tau_k}$ is the top-performing variant in CK+ and performs well in NEFER with LSTM, we will focus on this variant in subsequent experiments for clarity, unless otherwise specified.

\begin{table}[!h]
\centering
\footnotesize
\caption{Performance of the proposed event-based facial expression recognition framework evTransFER on the {\bf e-CK+} dataset,  CNN (U-Net) {\bf without} and {\bf with LSTM}, varying the temporal domain measurement and kernel function.
The best results for each variant are shown in bold letters. 
}
\footnotesize
\begin{tabular}{|l||cccc|}
\hline
\multirow{2}{*}{Encoder (Feat. Ext.)} & 
\multicolumn{4}{c|}{Top-1 Accuracy}                                                               \\ \cline{2-5} 
                                 &   \multicolumn{1}{c|}{$f_{\tau_k} \cdot h_{\tau_k}$}    & \multicolumn{1}{c|}{$f_{\tau_k} \cdot h_{\hat{\tau}_k}$} & \multicolumn{1}{c|}{$f_{\hat{\tau}_k} \cdot h_{\tau_k}$}        & $ f_{\hat{\tau}_k} \cdot h_{\hat{\tau}_k}$    \\ \hline \hline
evTransFER without LSTM &\multicolumn{1}{c|}{91,9\%} & \multicolumn{1}{c|}{90,5\%} & \multicolumn{1}{c|}{\textbf{92,4\%}} & 92,0\% \\ \hline
evTransFER with LSTM & 
\multicolumn{1}{c|}{93,0\%} & \multicolumn{1}{c|}{89,2\%} & \multicolumn{1}{c|}{\textbf{93,6\%}} & 93,1\% \\ \hline
\end{tabular}
\label{tab:TIE-performance-0}
\end{table}

\begin{table}[!h]
\centering
\footnotesize
\caption{Performance of the proposed event-based facial expression recognition framework evTransFER on the {\bf NEFER} dataset, CNN (U-Net) {\bf without} and {\bf with LSTM}, varying the temporal domain measurement and kernel function.
The best results for each variant are shown in bold letters.
}
\begin{tabular}{|l||cccc|}
\hline
\multirow{2}{*}{Encoder (Feat. Ext.)} & 
\multicolumn{4}{c|}{Top-1 Accuracy}                                                               \\ \cline{2-5} 
                                 &   \multicolumn{1}{c|}{$f_{\tau_k} \cdot h_{\tau_k}$}    & \multicolumn{1}{c|}{$f_{\tau_k} \cdot h_{\hat{\tau}_k}$} & \multicolumn{1}{c|}{$f_{\hat{\tau}_k} \cdot h_{\tau_k}$}        & $ f_{\hat{\tau}_k} \cdot h_{\hat{\tau}_k}$    \\ \hline \hline

 evTransFER without LSTM &  \multicolumn{1}{c|}{70,2\%} & \multicolumn{1}{c|}{76,5\%} & \multicolumn{1}{c|}{72,0\%} & \textbf{76,7\%} \\ \hline

evTransFER with LSTM&
\multicolumn{1}{c|}{71,4\%} & \multicolumn{1}{c|}{\textbf{76,0\%}} & \multicolumn{1}{c|}{75,5\%} & 75,9\% \\ \hline
\end{tabular}
\label{tab:TIE-performance-NEFER}
\end{table}

\subsection{Comparison with State-of-the-Art on the CK+ dataset}
\label{sec:sotacom}

Using the e-CK+ database and the setup described in Section~\ref{sec:evaluationSetUp}, in this section, we compare evTransFER with state-of-the-art networks for similar problems, for which the source code is publicly available. In this analysis, we included EST and Asynet, along with a Vision Transformer (ViT) baseline.

For EST, we use the original pipeline described by the authors in \cite{gehrig2019est}, with the voxel grid without normalization as input to a classification model, employing a ResNet-34 backbone with a pre-trained encoder fine-tuned for object classification. 
We also test the case in which three LSTM units are added after the backbone (with the encoder fixed) for EST, and before a fully connected layer. For Asynet, two methods described by the authors in \cite{messikommer2020asynet} are evaluated: \texttt{SSR + CNN}, where a traditional CNN is used for the sparse representation, and \texttt{SSR + SCC}, where a Submanifold Sparse Convolutional network is used, at a lower computational cost and better model performance. In both cases, the pipeline involves fine-tuning a pre-trained encoder for object classification, with VGG-13 as the backbone. For the ViT \cite{dosovitskiy2020image} we use the vit\_b\_16 model available in TorchVision. The training parameters for Asynet, EST, and ViT are given in Table~\ref{tab:methods_params}.

\begin{table}[!h]
\footnotesize
\centering
\caption{Performance on the \textbf{e-CK+ dataset} of state-of-the-art methods (SSR \& EST) and the here proposed method in facial expression recognition. All methods were trained and evaluated in this database. 
Asynet, EST, and ViT were trained using the original implementation (FT: fine-tuned). In evTransFER, the U-Net encoder is trained for face reconstruction (FR) and then transferred and fine-tuned to the proposed architecture. The best results with and without LSTM are shown in bold letters.} 
\label{tab:Methods-performance}
\begin{tabular}{|l|c|c|c|c|c|c|}
\hline
\multirow{2}{*}{Method / Reference} & 
\multicolumn{3}{c|}{Architecture} &
\multirow{2}{*}{\begin{tabular}[c]{@{}c@{}}Encoder\\ training\end{tabular}} & 
\multirow{3}{*}{\begin{tabular}[c]{@{}c@{}}Top-1\\ accuracy \\ (\%)\end{tabular}} & 
\multirow{3}{*}{\begin{tabular}[c]{@{}c@{}}Time\\ window\\ (ms)\end{tabular}} \\
\cline{2-4} & 
\begin{tabular}[c]{@{}c@{}}Event Data\end{tabular} & 
\begin{tabular}[c]{@{}c@{}}Encoder\\ (Feat. Ext.)\end{tabular} & 
\begin{tabular}[c]{@{}c@{}}Sequence\\ learning\end{tabular} & 
& & \\
\hline \hline
\multicolumn{1}{|l|}{\begin{tabular}[l]{@{}l@{}}Asynet CNN /      
\cite{messikommer2020asynet}\end{tabular}}                           
                         & \multicolumn{1}{c|}{SSR}                                                            & \multicolumn{1}{c|}{\begin{tabular}[c]{@{}c@{}} VGG-16\end{tabular}}         & -                                                           & FT                                                                  & 67,2                    & 33,3                                                             \\
\hline

\multicolumn{1}{|l|}{\begin{tabular}[l]{@{}l@{}}Asynet SSC /      
\cite{messikommer2020asynet}\end{tabular}}                   & \multicolumn{1}{c|}{SSR}                                                            & \multicolumn{1}{c|}{\begin{tabular}[c]{@{}c@{}}VGG-16\end{tabular}}         & -                                                           & FT                                                                  & 67,5                    & 33,3                                                                 \\ \hline

\multicolumn{1}{|l|}{\begin{tabular}[l]{@{}l@{}}EST+CNN /  \cite{gehrig2019est}\end{tabular}}                          & \multicolumn{1}{c|}{EST}                                                            & \multicolumn{1}{c|}{\begin{tabular}[c]{@{}c@{}}ResNet-34\end{tabular}}      & -                                                           &FT                                                                  & 70,9                    & 33,3                                                                 \\ \hline
\multicolumn{1}{|l|}{\begin{tabular}[l]{@{}l@{}}ViT /  \cite{dosovitskiy2020image}\end{tabular}}                          & \multicolumn{1}{c|}{TIE}                                                            & \multicolumn{1}{c|}{\begin{tabular}[c]{@{}c@{}} vit\_b\_16 \end{tabular}}      & -                                                           & FT                                                                  & 79.0                    & 33,3                                                                 \\ \hline

{evTransFER (ours)}                      & \multicolumn{1}{c|}{TIE}                                                            & \multicolumn{1}{c|}{\begin{tabular}[c]{@{}c@{}}U-Net\end{tabular}}          & -                                                           & FR \& FT     & {\bf 92,4}                    & 33,3                                                                 \\ \hline
\hline
\multicolumn{1}{|l|}{\begin{tabular}[l]{@{}l@{}}EST+CNN /  \cite{gehrig2019est}\end{tabular}}                            & \multicolumn{1}{c|}{EST}                                                            & \multicolumn{1}{c|}{\begin{tabular}[c]{@{}c@{}}ResNet-3\end{tabular}}      & LSTM                                                        & FT                                                                  & 83,5                    & 3 x 33,3                                                             \\ \hline
\multicolumn{1}{|l|}{\begin{tabular}[l]{@{}l@{}}ViT /  \cite{dosovitskiy2020image}\end{tabular}}                          & \multicolumn{1}{c|}{TIE}                                                            & \multicolumn{1}{c|}{\begin{tabular}[c]{@{}c@{}} vit\_b\_16 \end{tabular}}      & LSTM                                                           & FT                                                                  & 78.9                    & 3 x 33,3                                                                 \\ \hline

{{evTransFER (ours)}}                         & \multicolumn{1}{c|}{TIE}                                                            & \multicolumn{1}{c|}{\begin{tabular}[c]{@{}c@{}}U-Net\end{tabular}}          & LSTM                                                        & FR \& FT     & {\bf 93,6}                    & 3 x 33,3                                                             \\ \hline
\end{tabular}
\end{table}

Table~\ref{tab:Methods-performance} shows that our proposed method performs much better in the event-based facial expression recognition task (93,6\% of accuracy, with LSTM), over state-of-the-art methods. This corresponds to achieving 26.1\% points over SSR+SSC, 22.7\% points over the original EST (without LSTM), and 10.1\% over EST with LSTM.

In summary, the best results on the e-CK+ dataset are achieved using the proposed method evTransFER

\subsection{Ablation Study}
\label{sec:ablation}

To assess the significance of the proposed transfer learning methodology for training the U-Net Encoder in evTransFER, we compare three learning approaches: \begin{itemize}
\item {\bf training from scratch}, which involves no prior knowledge from facial frame reconstruction; 
\item {\bf proposed transfer learning without fine-tuning}, where the encoder is used with fixed weights from the facial frame reconstruction system; and 
\item {\bf proposed transfer learning with fine-tuning}, which involves using the facial frame reconstruction system and subsequently fine-tuning the encoder's weights.
\end{itemize}.
In all three approaches, we evaluate scenarios both with and without LSTM and consider the four TIE variants to clearly demonstrate the impact of the proposed transfer learning approach.

Table~\ref{tab:TIE-performance-I} presents the results for the proposed method variants with and without LSTM on the e-CK+ dataset. The training approach for the encoder influences the accuracy of the facial expression classifier (Table~\ref{tab:TIE-performance-I}). Training from scratch, without prior knowledge of event-based facial analysis components, yields average performance (76.7\% in $f_{\hat{\tau}_k} \cdot h_{\hat{\tau}_k}$ without LSTM; and 80.6\% in $f_{\hat{\tau}_k} \cdot h_{\hat{\tau}_k}$ with LSTM). Using the reconstruction generator network encoder weights as a static feature extractor improves accuracy by approximately 10\% (86.6\% in $f_{\hat{\tau}_k} \cdot h_{\tau_k}$ without LSTM; and 88.7\% in $f_{\hat{\tau}_k} \cdot h_{\tau_k}$ with LSTM). When fine-tuning by retraining all network weights, including the transferred encoder trained for face reconstruction, the model achieves higher accuracy (92.4\% in $f_{\hat{\tau}_k} \cdot h_{\tau_k}$ without LSTM; and 93.6\% in $f_{\hat{\tau}_k} \cdot h_{\tau_k}$ with LSTM) compared to the other scenarios.


\begin{table}[!ht]
\centering
\footnotesize
\caption{Performance of the proposed event-based facial expression recognition framework evTransFER on the e-CK+ database, varying the temporal domain, measurement function, and kernel function, and {\bf applying different encoder training techniques}. The best results for each variant are shown in bold letters. 
}
\begin{tabular}{|l||l|cccc|}
\hline
\multirow{2}{*}{Encoder (Feat. Ext.)} & \multirow{2}{*}{Encoder training} & \multicolumn{4}{c|}{Top-1 Accuracy  {\bf without LSTM} (e-CK+)}                                                               \\ \cline{3-6} 
                                 &&  \multicolumn{1}{c|}{$f_{\tau_k} \cdot h_{\tau_k}$}    & \multicolumn{1}{c|}{$f_{\tau_k} \cdot h_{\hat{\tau}_k}$} & \multicolumn{1}{c|}{$f_{\hat{\tau}_k} \cdot h_{\tau_k}$}        & $ f_{\hat{\tau}_k} \cdot h_{\hat{\tau}_k}$    \\ \hline \hline
evTransFER: CNN (U-Net) & trained from scratch                   & \multicolumn{1}{c|}{72.5\%} & \multicolumn{1}{c|}{73.9\%} & \multicolumn{1}{c|}{71.8\%} & \textbf{76.4}\% \\ \hline
evTransFER: CNN (U-Net) & transferred               & \multicolumn{1}{c|}{83,1\%} & \multicolumn{1}{c|}{85,8\%} & \multicolumn{1}{c|}{\textbf{86,6\%}} & 85,7\% \\ \hline
evTransFER: CNN (U-Net) & transferred \&  fine-tuned             & \multicolumn{1}{c|}{91,9\%} & \multicolumn{1}{c|}{90,5\%} & \multicolumn{1}{c|}{\textbf{92,4\%}} & 92,0\% \\ \hline
\end{tabular}
\label{tab:TIE-performance-I}

\vspace{2.5mm}

\begin{tabular}{|l||l|cccc|}
\hline
\multirow{2}{*}{Encoder (Feat. Ext.)} & \multirow{2}{*}{Encoder training} & \multicolumn{4}{c|}{Top-1 Accuracy  {\bf with LSTM} (e-CK+)}                                                               \\ \cline{3-6} 
                                 &&   \multicolumn{1}{c|}{$f_{\tau_k} \cdot h_{\tau_k}$}    & \multicolumn{1}{c|}{$f_{\tau_k} \cdot h_{\hat{\tau}_k}$} & \multicolumn{1}{c|}{$f_{\hat{\tau}_k} \cdot h_{\tau_k}$}        & $ f_{\hat{\tau}_k} \cdot h_{\hat{\tau}_k}$    \\ \hline \hline
evTransFER: CNN (U-Net) & trained from scratch                   & \multicolumn{1}{c|}{70.2\%} & \multicolumn{1}{c|}{76.9\%} & \multicolumn{1}{c|}{74.3\%} & \textbf{80.6\%} \\ \hline
evTransFER: CNN (U-Net) & transferred              & \multicolumn{1}{c|}{86,9\%} & \multicolumn{1}{c|}{88,3\%} & \multicolumn{1}{c|}{\textbf{88,7\%}} & 88,2\% \\ \hline
evTransFER: CNN (U-Net) & transferred \&  fine-tuned &
\multicolumn{1}{c|}{93,0\%} & \multicolumn{1}{c|}{89,2\%} & \multicolumn{1}{c|}{\textbf{93,6\%}} & 93,1\% \\ \hline
\end{tabular}
\label{tab:TIE-performance-II}
\end{table}


\begin{table}[!ht]
\centering
\footnotesize
\caption{Performance of the proposed event-based facial expression recognition framework evTransFER on the NEFER database, varying the temporal domain, measurement function, and kernel function, and {\bf applying different encoder training techniques}. The best results for each variant are shown in bold letters.}
\begin{tabular}{|l||l|cccc|}
\hline
\multirow{2}{*}{Encoder (Feat. Ext.)} & \multirow{2}{*}{Encoder training} & \multicolumn{4}{c|}{Top-1 Accuracy {\bf without LSTM} (NEFER)} \\ \cline{3-6} 
 & & \multicolumn{1}{c|}{$f_{\tau_k} \cdot h_{\tau_k}$} & \multicolumn{1}{c|}{$f_{\tau_k} \cdot h_{\hat{\tau}_k}$} & \multicolumn{1}{c|}{$f_{\hat{\tau}_k} \cdot h_{\tau_k}$} & $f_{\hat{\tau}_k} \cdot h_{\hat{\tau}_k}$ \\ \hline \hline
 
evTransFER: CNN (U-Net) & trained from scratch       & \multicolumn{1}{c|}{43,9\%} & \multicolumn{1}{c|}{45,5\%} & \multicolumn{1}{c|}{47,3\%} & \textbf{48,8\%} \\ \hline

evTransFER: CNN (U-Net) & transferred                & \multicolumn{1}{c|}{70,6\%} & \multicolumn{1}{c|}{\textbf{73,8\%}} & \multicolumn{1}{c|}{71,5\%} & 67,2\% \\ \hline

evTransFER: CNN (U-Net) & transferred \& fine-tuned  & \multicolumn{1}{c|}{70,2\%} & \multicolumn{1}{c|}{76,5\%} & \multicolumn{1}{c|}{72,0\%} & \textbf{76,7\%} \\ \hline

\end{tabular}
\label{tab:NEFER-TIE-performance}

\vspace{2.5mm}

\begin{tabular}{|l||l|cccc|}
\hline
\multirow{2}{*}{Encoder (Feat. Ext.)} & \multirow{2}{*}{Encoder training} & \multicolumn{4}{c|}{Top-1 Accuracy {\bf with LSTM} (NEFER)} \\ \cline{3-6} 
 & & \multicolumn{1}{c|}{$f_{\tau_k} \cdot h_{\tau_k}$} & \multicolumn{1}{c|}{$f_{\tau_k} \cdot h_{\hat{\tau}_k}$} & \multicolumn{1}{c|}{$f_{\hat{\tau}_k} \cdot h_{\tau_k}$} & $f_{\hat{\tau}_k} \cdot h_{\hat{\tau}_k}$ \\ \hline \hline

evTransFER: CNN (U-Net) & trained from scratch       & 
\multicolumn{1}{c|}{43,3\%} & \multicolumn{1}{c|}{47,1\%} & \multicolumn{1}{c|}{45,0\%} & \textbf{49,7\%} \\ \hline

evTransFER: CNN (U-Net) & transferred                & 
\multicolumn{1}{c|}{64,0\%} & \multicolumn{1}{c|}{72,0\%} & \multicolumn{1}{c|}{70,3\%} & \textbf{74,1\%} \\ \hline

evTransFER: CNN (U-Net) & transferred \& fine-tuned  & 
\multicolumn{1}{c|}{71,4\%} & \multicolumn{1}{c|}{\textbf{76,0\%}} & \multicolumn{1}{c|}{75,5\%} & 75,9\% \\ \hline

\end{tabular}
\end{table}

A complementary ablation analysis was conducted on the NEFER dataset (Table~\ref{tab:NEFER-TIE-performance}). For real event data, accuracy values are lower than on the e-CK+ dataset due to increased variability, sparser events, and sensor noise in NEFER recordings. The performance patterns remain consistent across datasets. Training the encoder from scratch yields limited recognition performance (43\% to 49\%, depending on TIE representation and LSTM use), confirming that learning event-based facial representations directly from NEFER is challenging when samples show low event density. Using the reconstruction-trained encoder as a fixed feature extractor improves performance to 73.8\% without LSTM and 74.1\% with LSTM - a gain of over 25 percentage points compared to training from scratch. This underscores the importance of transferring facial structure priors learned during reconstruction. With fine-tuning, additional gains occur, especially for the $f_{\tau_k} \cdot h_{\hat{\tau}_k}$ and $f_{\hat{\tau}_k} \cdot h_{\hat{\tau}_k}$ variants. The results of 76.7\% (without LSTM) and 76.0\% (with LSTM) confirm that adapting the pretrained encoder to NEFER events provides benefits.
Unlike in e-CK+, the NEFER dataset shows that LSTM does not always improve performance.

\begin{figure}[!ht]
    \centering
    \subcaptionbox{evTransFER: TIE + CNN (U-Net): Encoder trained from scratch - $f_{\hat{\tau}_k} \cdot h_{\tau_k}$. Accuracy: 76.4\%} 
    [0.49\textwidth]{\includegraphics[width=0.9\linewidth]{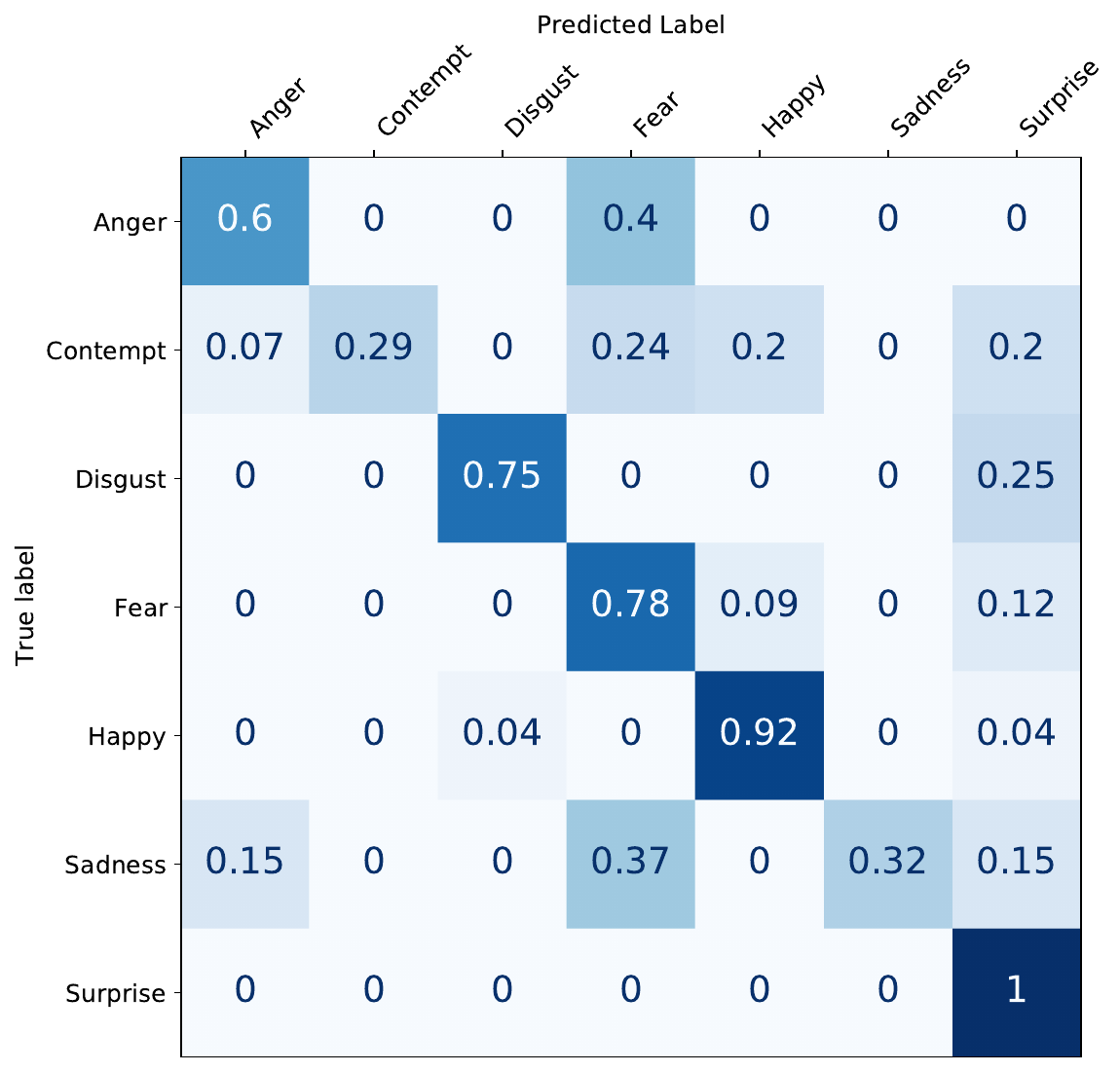 }}
    \subcaptionbox{evTransFER: TIE + CNN (U-Net): Encoder transferred from FR - $f_{\hat{\tau}_k} \cdot h_{\tau_k}$. Accuracy: 86.7\%} 
    [0.49\textwidth]{\includegraphics[width=0.9\linewidth]{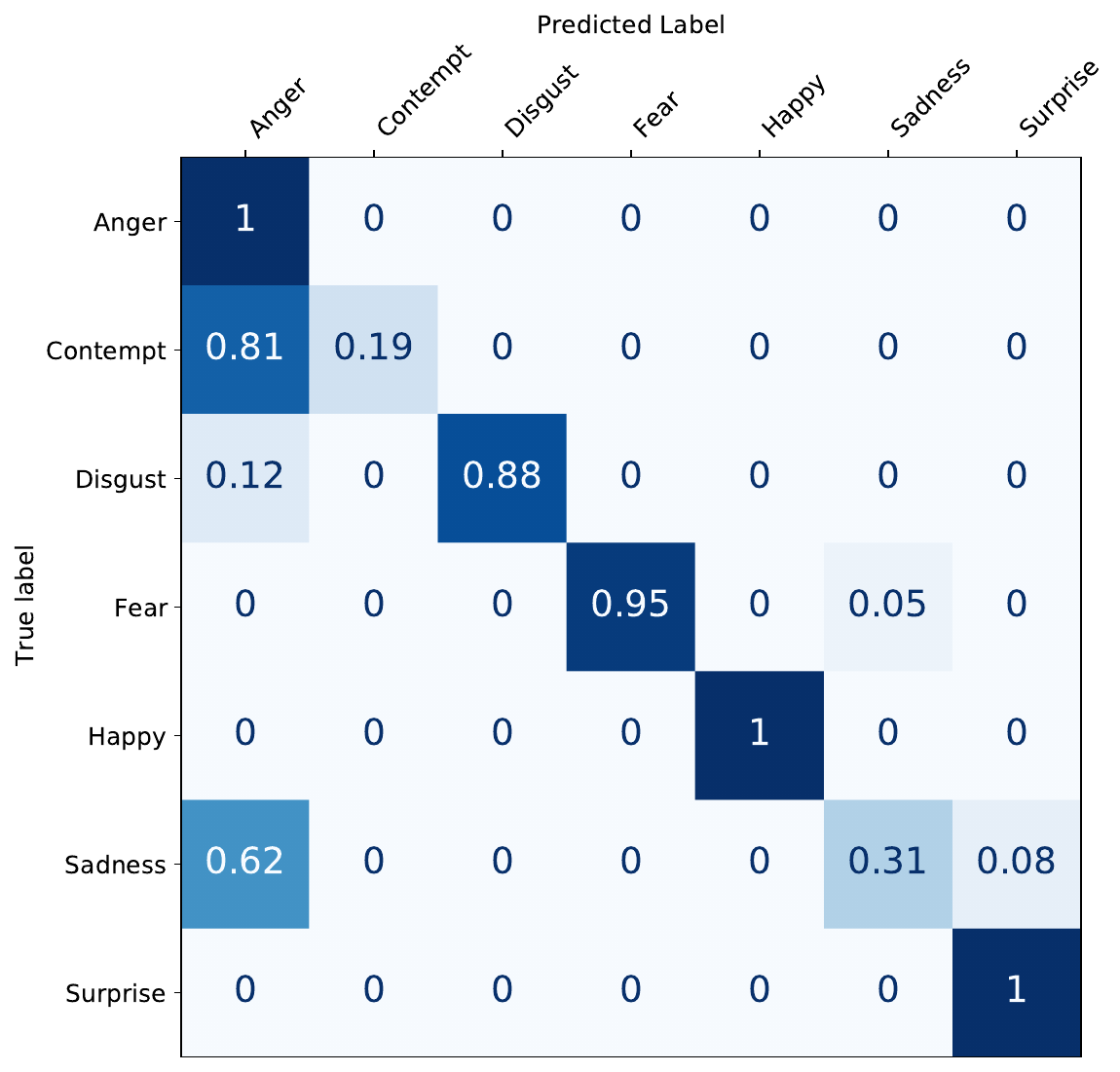}}
    \subcaptionbox{evTransFER: TIE + CNN (U-Net): Encoder transferred from FR and fine-tuned - $f_{\hat{\tau}_k} \cdot h_{\tau_k}$. Accuracy: 92.4\%} 
    [0.49\textwidth]{\includegraphics[width=0.9\linewidth]{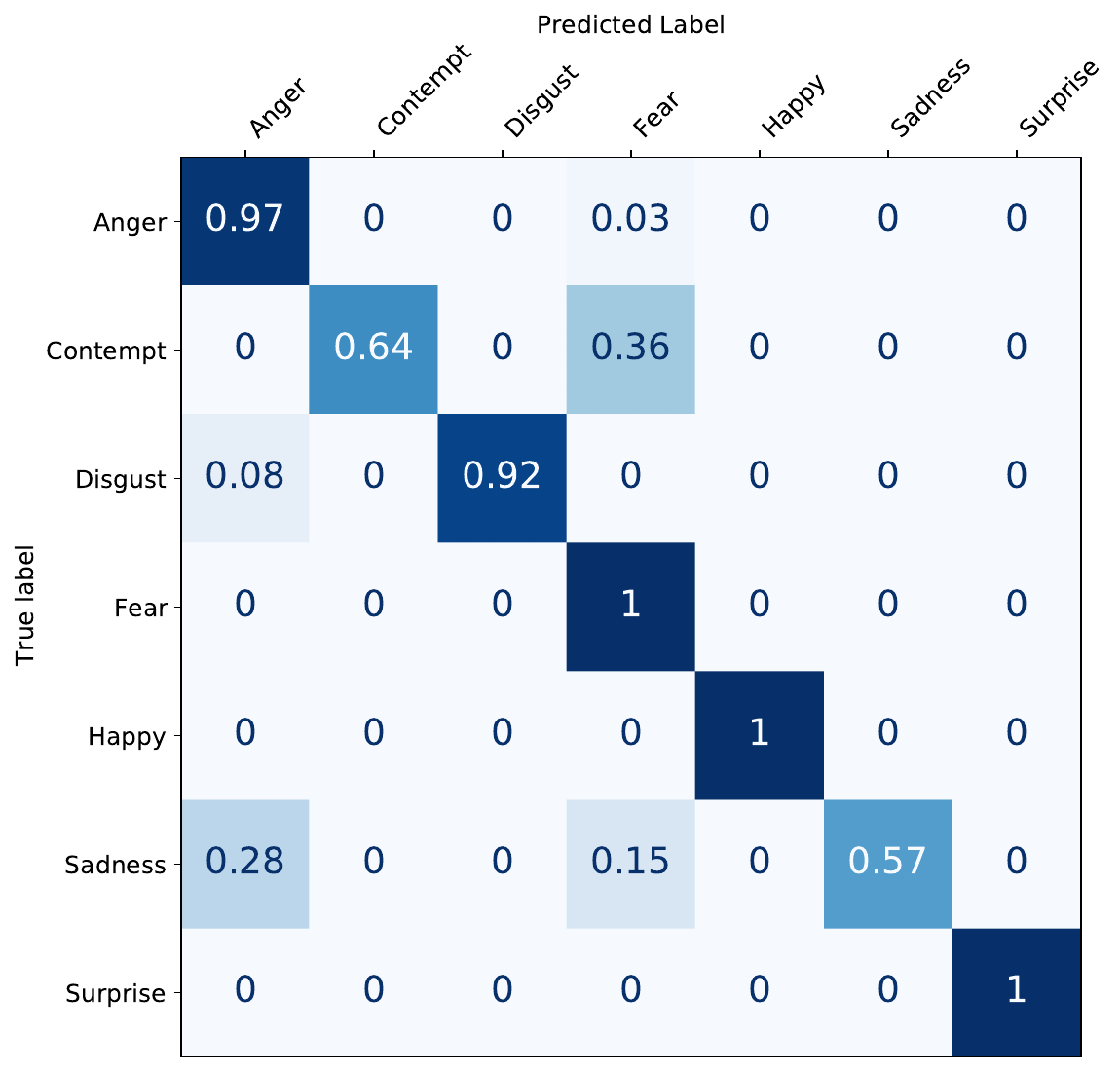}}
\caption{Confusion Matrices of the proposed event-based facial expression recognition framework evTransFER ($f_{\hat{\tau}_k} \cdot h_{\tau_k}$) and ablations.
Left: without LSTM. Right: with LSTM. Top: Encoder Trained from scratch. Center: Encoder Trained for Face Recognition. Bottom: Encoder Trained for Face Recognition and then fine-tuned.}
\label{fig:conf_matrices_1}
\end{figure}

To better illustrate the effect of each training procedure and the use of LSTM, the analysis is complemented by the confusion matrices presented in Figure~\ref{fig:conf_matrices_1}, where we report the best results achieved by the TIE representation ($f_{\tau_k} \cdot h_{\tau_k}$) on the CK+ dataset. As can be observed, the TIE + CNN (ResNet-34) model trained from scratch shows inconsistent results across the different classes (Figure~\ref{fig:conf_matrices_1} (a)), especially with false positives and negatives in the ``Fear" and ``Anger" classes, low accuracy for the classes Contempt and Sadness, while classes such as ``Happy" and ``Surprise" show good results. 

\subsection{Training and Processing times}

We used an NVIDIA DGX-1, a purpose-built system optimized for deep learning. The specifications are presented in Table~\ref{tab:DGX-1}, where up to 2 GPU cores were used for event-based database generation (1 week for e-CK+), event-based encoder training via face reconstruction proxy ($\approx10$ hours of training, with 100 epochs), and the application of the classifier training techniques (a few days).

\begin{table}[H]
\centering
\footnotesize
\caption{Hardware specifications used for experiments and tests in event-based facial expression recognition.}
\footnotesize
\begin{tabular}{|c|c|c|c|c|c|c|c|}
\hline
Hardware     & Accelerator & Boost clock & Memory Clock & Bandwidth & VRAM & GPU\\ \hline \hline
NVIDIA DGX-1 & V100        & 1530 MHz  & 1.75Gbit/s   & 900GB/sec       & 32GB & GV100 \\ \hline
\end{tabular}
\label{tab:DGX-1}
\end{table}

Regarding the inference computation times, we used an NVIDIA GeForce GTX 1050 4GB, with the results for the different methods listed in Table~\ref{tab:inference-time}. We observe that the proposed method, evTransFER, takes only 3.7 ms (1.6 ms more than EST without LSTM), and when combined with an LSTM, it takes a time comparable to Asynet.
Thus, there is a trade-off: incorporating LSTM temporal models with better results implies a larger sample time window, in this particular case, from 33.3 ms (in the architecture without LSTM) to 3x33 ms (in the architecture with LSTM), which increases processing time and has high latency, but still allows the system to run at the equivalent of 33fps (due to the event-windows size) and with a latency of approximately 25ms.

\begin{table}[H]
\centering
\footnotesize
\caption{Inference times for event-based facial expression recognition methods.}
\footnotesize
\begin{tabular}{|c|c|c|c|}
\hline
EST (without LSTM) & Asynet & evTransFER (without LSTM) & evTransFER (with LSTM)\\ \hline \hline
2,09 [ms] & 23,4 [ms] & 3,71 [ms] & 24,69 [ms] \\ \hline
\end{tabular}
\label{tab:inference-time}
\end{table}

\section{Discussion}
\label{sec:discussion}

\subsection{Indirect Comparison on various Datasets}

Table~\ref{tab:sota-fer} summarizes performance reported in recent literature using datasets other than e-CK+ employed here. Although a direct comparison between these results and evTransFER is not feasible because of dataset differences and preprocessing, our proposal evTransFER (TIE + CNN trained for FR and fine-tuned + LSTM) outperforms existing methods with 93.6\% accuracy. We use the e-CK+ dataset with 346x260 spatial resolution. Two studies report results using similar datasets: ANN+ResNet-18 \cite{barchid2024spikingfer} achieves 92.0\% on DVS-CK+ (200x200 pixels), and EST+CNN+LSTM \cite{verschae2023eventgesturefer} reports 89.1\% on e-CK+ (640x480 pixels). \cite{barchid2024spikingfer} preprocesses faces by aligning and cropping them to enhance performance, whereas \cite{verschae2023eventgesturefer} shows that higher-resolution sensors improve outcomes (results shown at 640x480 for face expression recognition with sequence learning). We use reduced resolution of 346x260 for e-CK+ compared to \cite{verschae2023eventgesturefer}, without adjusting images through face alignment or cropping as in \cite{barchid2024spikingfer}. 

\begin{table}[!h]
\centering
\footnotesize
\caption{Reported performance of state-of-the-art methods on the DVS-CK+, DVS-ADFES, DVS-CASIA, DVS-MMI, NEFER, e-CK+ and e-MMI datasets. The methods use datasets with different spatial resolutions. FR: face reconstruction; FT: fine-tuning.}
\label{tab:sota-fer}
\begin{tabular}{|l||c|c|c|c|c|c|}
\hline
\begin{tabular}[c]{@{}l@{}}Database \\ / Reference\end{tabular}  
& Method 
& \begin{tabular}[c]{@{}c@{}}Event\\ Data\end{tabular} 
& \begin{tabular}[c]{@{}c@{}}Encoder\\ training\end{tabular} 
& \begin{tabular}[c]{@{}c@{}}Top-1\\ accuracy\end{tabular} 
& \begin{tabular}[c]{@{}c@{}}Time\\ window\end{tabular} 
& \begin{tabular}[c]{@{}c@{}}Sensor \\ Spatial\\ Resolution\end{tabular} \\ \hline \hline

DVS-CK+ / \cite{barchid2024spikingfer}   
& ANN                     
& Frames                                         
& \begin{tabular}[c]{@{}l@{}}From \\ Scratch\end{tabular}
& 92,0\%                    
& 6 x 33ms 
& 200x200 \\ \hline

DVS-CK+ / \cite{barchid2024spikingfer}   
& SNN                     
& Tensor                                         
& \begin{tabular}[c]{@{}l@{}}From \\ Scratch\end{tabular}
& 89,3\%                    
& 6 x 33ms  
& 200x200 \\ \hline

DVS-ADFES / \cite{barchid2024spikingfer} 
& ANN                     
& Frames                                         
& \begin{tabular}[c]{@{}l@{}}From \\ Scratch\end{tabular}
& 79,6\%                    
& 6 x 33ms    
& 200x200 \\ \hline

DVS-ADFES / \cite{barchid2024spikingfer} 
& SNN                     
& Tensor                                         
& \begin{tabular}[c]{@{}l@{}}From \\ Scratch\end{tabular}
& 74,2\%                    
& 6 x 33ms   
& 200x200 \\ \hline

DVS-CASIA / \cite{barchid2024spikingfer} 
& ANN                     
& Frames                                         
& \begin{tabular}[c]{@{}l@{}}From \\ Scratch\end{tabular}
& 72,4\%                    
& 6 x 33ms 
& 200x200 \\ \hline

DVS-CASIA / \cite{barchid2024spikingfer} 
& SNN                     
& Tensor                                         
& \begin{tabular}[c]{@{}l@{}}From \\ Scratch\end{tabular}
& 69,7\%                    
& 6 x 33ms  
& 200x200 \\ \hline

DVS-MMI / \cite{barchid2024spikingfer}   
& ANN                     
& Frames                                         
& \begin{tabular}[c]{@{}l@{}}From \\ Scratch\end{tabular}
& 62,0\%                    
& 6 x 33ms 
& 200x200 \\ \hline

DVS-MMI / \cite{barchid2024spikingfer}   
& SNN                     
& Tensor                                         
& \begin{tabular}[c]{@{}l@{}}From \\ Scratch\end{tabular}
& 63,1\%                    
& 6 x 33ms  
& 200x200 \\ \hline

NEFER / \cite{berlincioni2023nefer}                     
& CNN                     
& TBR \cite{innocenti2021tbr}                                                 
& \begin{tabular}[c]{@{}l@{}}From \\ Scratch\end{tabular}
& 31,0\%                    
& 15ms 
& 1280x720 \\ \hline

\begin{tabular}[c]{@{}l@{}}\textbf{NEFER /}\\ \textbf{evTransFER}\end{tabular}
& \textbf{CNN+LSTM}  
& TIE 
& \begin{tabular}[c]{@{}l@{}}FR \& FT\end{tabular} 
& \textbf{76,7\%} 
& 3 x 33ms 
& 1280x720 \\ \hline

e-CK+ / \cite{verschae2023eventgesturefer}                     
& CNN+LSTM                
& EST \cite{gehrig2019est}                                                  
& \begin{tabular}[c]{@{}l@{}}From \\ Scratch\end{tabular}
& 89,1\%                    
& 3 x 33ms 
& 640x480 \\ \hline

e-MMI / \cite{verschae2023eventgesturefer}                      
& CNN+LSTM                
& EST \cite{gehrig2019est}                                                  
& \begin{tabular}[c]{@{}l@{}}From \\ Scratch\end{tabular}
& 83,7\%                    
& 3 x 33ms 
& 640x480 \\ \hline

\begin{tabular}[c]{@{}l@{}}\textbf{e-CK+ /}\\ \textbf{evTransFER}\end{tabular}
& \textbf{CNN+LSTM}  
& TIE 
& \begin{tabular}[c]{@{}l@{}}FR \& FT\end{tabular} 
& \textbf{93,6\%} 
& 3 x 33ms 
& 346x260 \\ \hline
\end{tabular}
\end{table}

\subsection{TIE representation and Long-term modeling}

The experimental analysis highlights the importance of event-based representation and training techniques. The measurement function in TIE plays a crucial role, often surpassing the kernel in classifier performance. The findings emphasize the importance of selecting appropriate event representations, effective network architectures, and suitable training techniques for event-based facial expression recognition.

We demonstrate that incorporating LSTM memory units allows evTransFER to sequentially learn from longer-term information on CK+, although no improvement is observed in NEFER. Adding LSTM units requires more processing time but enhances accuracy. Using a 33ms event window (without an LSTM) enables recognition at 33~fps with a processing time of 3.71ms, introducing minimal latency. The systems could be used in real-time. With an LSTM, the latency is 24.69ms, yet the system operates at 33~fps by considering the current 33ms window and two previous ones. This latency could be reduced by reusing the TIE representation and Feature Extraction from previous windows.

\subsection{Reconstruction as a Proxy for Encoder Training}

A transfer-learning technique enhances framework performance by leveraging a U-Net-based encoder trained for face reconstruction from event sequences. This  learning approach requires, during training,only events and corresponding images of the same face, without additional annotations. The reconstruction weights are fine-tuned within the classification system, improving recognition accuracy by 21.5 percentage points, from 72.1\% to 92.4\% on CK+. We believe this improvement stems from the U-Net's ability to learn spatio-temporal dynamics in facial expressions, which are crucial for facial expression recognition.

Prior knowledge from facial frame reconstruction significantly enhances model accuracy, as shown in the confusion matrices. When trained from scratch, the encoder's performance on two of the face expression classes is poor, with an accuracy of approximately 30\%. However, training the encoder for recognition and fine-tuning leads to improvements across classes, with the lowest-performing classes reaching 60\% and 72\% accuracy, and others exceeding 93\%. Supervision during encoder reconstruction training helps to understand the spatio-temporal dynamics of facial expressions. This, combined with fine-tuned supervision and an LSTM, effectively captures the structure of facial expression recognition.

We emphasize employing reconstruction as a proxy for training the encoder as a feature extractor in event-based cameras. To understand this, it is important to recognize that event-based datasets remain limited, with fewer real-world datasets containing annotated data. Currently, systems can generate datasets from videos and simulators. Additionally, cameras like DAVIS from iniVation capture event sequences with corresponding frames. By training the encoder through reconstruction as a proxy, along with existing data systems, we can develop encoders without dataset annotations. The image frame to be reconstructed serves as the desired outcome and is required only during training. This strategy could be applied to construct encoders for other object classification problems.

\section{Conclusions}
\label{sec:conclusions}

This paper introduces a novel learning and classification framework for event-based facial expression recognition, termed evTransFER. This framework features TIE, a representation that effectively normalizes the temporal variable within the structure's measurement function, enabling a more precise characterization of the spatio-temporal distribution of events. Additionally, it incorporates LSTM to capture long-term dynamics. Furthermore, evTransFER employs a transfer-learning technique utilizing a U-Net-based Encoder, transferred from a facial reconstruction network, significantly enhancing the performance of our recognition framework and offering potential applicability in similar classification contexts.

We assessed the proposed framework using the e-CK+ database. In a state-of-the-art comparative evaluation,evTransFER outperformed other event-based classification models in facial expression recognition. By employing the proposed TIE representation as a latent variable for both reconstruction and classification, the proposed method achieved accuracies of 92.4\% without LSTM and 93.6\% with LSTM-based temporal learning. Compared to state-of-the-art models, evTransFER significantly outperforms them, achieving 26.1\% higher accuracy than SSR+SSC, 22.7\% higher than the original EST (without LSTM), and 10.1\% higher than EST with LSTM. This enhanced performance is attributed to our event-based facial expression recognition framework, which integrates transfer learning and fine-tuning from facial reconstruction, leveraging prior knowledge and learning the spatio-temporal dynamics of event sequences, as demonstrated in the ablation studies.

In addition to the synthetic e-CK+ dataset, we evaluated evTransFER on NEFER, a real neuromorphic facial expression database. This second evaluation shows that the proposed reconstruction-driven transfer learning strategy substantially improves recognition in real event data, increasing accuracy by more than 25 percentage points compared to training the encoder from scratch. Although the absolute performance on NEFER is naturally lower due to sensor noise, sparsity, and non-uniform event activity, the same patterns observed in e-CK+ are preserved: (i) transfer learning consistently boosts recognition performance, (ii) fine-tuning enables the model to better adapt to the statistics of real event sequence, and (iii) the temporal domain and kernel configuration $f_{\tau_k} \cdot h_{\hat{\tau}_k}$ remains the most reliable variant. On the other hand, in the NEFER dataset, the use of a LSTM did not improve results.

Future work will explore the use of the learned encoder for other problems (face detection, face recognition, etc.). We also plan to evaluate and adapt the proposed method to classification problems in which the spatio-temporal dynamics of other objects are key, and no annotated data are available; in particular, we will further explore the strategy of building encoders using frame reconstruction as a proxy.

\section*{Acknowledgments}
\label{sec:ack}

This work was partially funded by the FONDEQUIP Project EQM170041.


 \bibliographystyle{model1-num-names}

\bibliography{pr-refs}

\end{document}